\documentclass{article}
\usepackage{amsmath}

\PassOptionsToPackage{numbers, compress}{natbib}
\usepackage[eandd, final]{neurips_2026}

\usepackage[utf8]{inputenc} 
\usepackage[T1]{fontenc}    
\usepackage{hyperref}       
\usepackage{url}            
\usepackage{booktabs}       
\usepackage{amsfonts}       
\usepackage{nicefrac}       
\usepackage{microtype}      
\usepackage{xcolor}         

\usepackage{lipsum} 
\usepackage{bm}
\usepackage{graphicx}
\usepackage{pifont}
\newcommand{\cmark}{\ding{51}}
\newcommand{\xmark}{\ding{55}}
\usepackage{ulem}

\usepackage[font=small,labelfont=bf]{caption}

\usepackage{makecell}

\usepackage{etoc}

\usepackage{colortbl}

\usepackage{algorithm}
\usepackage{algpseudocode}
\usepackage{multirow}
\usepackage{subcaption}

\definecolor{goodgreen}{RGB}{198, 239, 206}
\definecolor{badred}{RGB}{255, 199, 206}
\definecolor{midyellow}{RGB}{255, 235, 156}
\definecolor{darkpurple}{RGB}{96,0,128}
\definecolor{brightpurple}{RGB}{178,50,200}    

\newcommand{\chg}[1]{#1}

\newcommand{\red}[1]{\textcolor{black}{#1}}

\definecolor{revold}{RGB}{0,140,60}

\definecolor{revnew}{RGB}{220,0,0}

\title{ForceBody: Force-Paired Parametric Body Motion with Torque Uncertainty}

\author{%
  \textbf{Joonwoo Kwon}$^{1}$ \quad
  \textbf{Yufei Zhang}$^{2}$ \quad
  \textbf{Xiaoming Liu}$^{3}$ \quad
  \textbf{Zijun Cui}$^{1}$ \\[0.5em]
  $^{1}$Michigan State University \qquad
  $^{2}$Independent Researcher \\
  $^{3}$University of North Carolina at Chapel Hill \\[0.3em]
  \texttt{\{kwonjoon;cuizijun\}@msu.edu, yufeizhang96@outlook.com, liuxm@cs.unc.edu}
}

\begin{document}
\raggedbottom
\maketitle

\begin{figure*}[h]
\vspace{-7mm}
\centering
\includegraphics[width=\dimexpr\textwidth-3pt\relax]{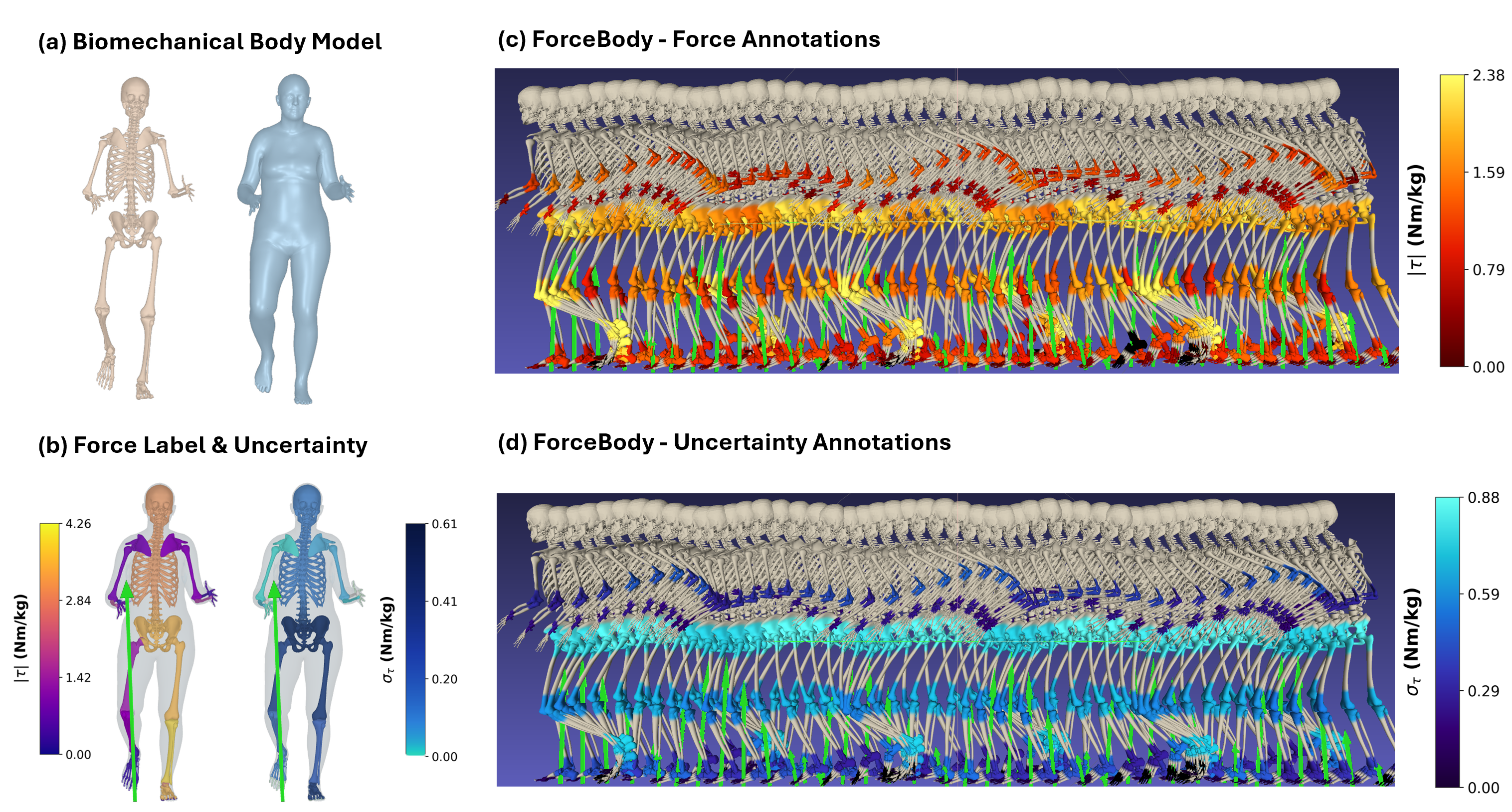}
\caption{\textbf{ForceBody Overview.}
\textbf{(a)} ForceBody adopts SKEL~\cite{keller2023skin} as the parametric
body model: a biomechanical skeleton (left) inside the SMPL surface (right).
\textbf{(b)} For one representative frame, the left body is colored by joint
torque $|\tau|$ (warm), the right by Monte Carlo uncertainty $\sigma_\tau$
(cool). Green arrows on the feet are measured GRF.
\textbf{(c)} Force annotations on a walking sequence; bones colored by
$|\tau|$ (Nm/kg, right colorbar), green arrows on the feet are measured GRF.
\textbf{(d)} Uncertainty annotations on the same sequence, bones colored
by $\sigma_\tau$ (Nm/kg). The body mesh in (c) and (d) is omitted for visualization clarity.
\vspace{-1mm}}
\label{fig:teaser}
\end{figure*}



\begin{abstract}
Forces and torques drive human motion, central to biomechanics, robotics, and physics-aware animation, yet they are often absent from learning-based motion pipelines for body reconstruction and generation. Some motion datasets provide force annotations, but the paired motion sequences are incompatible with the parametric body models used in modern learning pipelines. In addition, existing datasets typically obtain joint torques through inverse dynamics, which is inherently sensitive to the quality of the input kinematics; however, they do not quantify the reliability of these torque estimates. We close this gap with \textbf{ForceBody}, the first dataset to bring ground truth force supervision into a parametric body representation, shipped with per-sample torque uncertainty. ForceBody pairs the SKEL body model with ground reaction forces and inverse-dynamics joint torques across \red{$10{,}386$} motion trials ($9.7$M frames, $26.9$ hours) from $140$ subjects. Each torque label is paired with a per-frame, per-joint uncertainty annotation obtained by Monte Carlo sampling through the inverse dynamics pipeline. We benchmark six architectures (MLP, Conv1D, LSTM, GRU, Mamba2, Transformer) on ground reaction force and joint torque prediction from SKEL motion. As one example use of the released uncertainty, finetuning a Transformer with per-sample uncertainty weighting reduces the MAE of joint torque estimation by $5\%$. Code and data are at \href{https://github.com/DominoAI-Lab/ForceBody-NeurIPS-ED-2026}{https://github.com/DominoAI-Lab/ForceBody-NeurIPS-ED-2026}.
\end{abstract}


\section{Introduction}

Capturing human motion in 3D body pose and shape has become increasingly 
prevalent with recent advances in computer vision, computer graphics, 
and deep learning \cite{zhang2023body,tesch2025bedlam2,yang2026sam3dbody}.  
Nonetheless, the forces, including joint torques 
and ground reaction forces (GRF), that produce the motion are primarily absent from the estimation. 
These physical signals are critical to applications 
ranging from biomechanical assessment~\cite{holder2020systematic,
kalkhoven2021training} to exoskeleton and prosthetic 
design~\cite{molinaro2020biological} and sports performance 
analysis~\cite{johnson2018predicting}. 
Concurrent efforts tackle this limitation by incorporating dynamics into learning-based motion estimation, but progress is bottlenecked on two fronts. On the training side, dynamics is typically incorporated via external simulators rather than end-to-end learning. On the evaluation side, metrics rely on kinematic surrogate of physics plausibility, {\it e.g.}, motion smoothness~\cite{zhang2024physpt}, motion replay stability~\cite{zhao2025pp}, rather than the dynamic quantities themselves. Both limitations trace to the same root cause: real-world force annotations are unavailable in parametric body representations. Closing this gap requires a force-paired parametric body dataset, where forces are derived from measurements, in a form learning-based body motion pipelines can directly consume. 

The AddBiomechanics (AddB) dataset~\cite{werling2024addbiomechanics, werling2023addbiomechanics} is the largest publicly available collection of human motion paired with measured GRF and inverse-dynamics torques, aggregating 15 sources of motion-and-force data. Its motion, however, is provided as marker trajectories ({\it i.e.}, 3D positions of motion-capture markers placed on the body)
rather than the parametric body parameters ({\it e.g.}, SMPL~\cite{loper2023smpl}, SKEL~\cite{keller2023skin}) native to motion reconstruction and generation pipelines. Prior attempts to pair parametric body models with physical labels remain limited. GroundLink~\cite{han2023groundlink} provides force-plate-measured GRF alongside SMPL motion but is small in scale and does not include joint torques. ImDy~\cite{liu2025imdy} applies physics simulation to SMPL motion from AMASS~\cite{mahmood2019amass} to produce forces and torques at scale, but its labels are simulator outputs rather than physically measured. As a result, no existing dataset offers measured forces in a parametric body representation.

The reliability of torque annotations poses a further challenge. Joint torques cannot be directly measured. They are computed through inverse dynamics from motion kinematics and external forces such as GRF~\cite{winter2009biomechanics}. Inverse dynamics is inherently sensitive to kinematic noise: small positional perturbations at the input can propagate and yield large deviations in estimated torques~\cite{uchida2022conclusion, reinbolt2007patient, riemer2008uncertainties}. This uncertainty is intrinsic to any dataset that derives torques from inverse dynamics, AddBiomechanics \cite{werling2024addbiomechanics} included. Yet no existing force dataset releases per-sample torque reliability annotations.

\begin{table}[t!]
\centering
\footnotesize
\caption{Comparison of force-paired human motion datasets. $^*$UnderPressure provides vertical GRF only; \chg{$^\dagger$}VidSole reports the knee adduction moment only.}
\label{tab:dataset_comparison}
\setlength{\tabcolsep}{5pt}
\renewcommand{\arraystretch}{0.95}
\begin{tabular}{l| c| ccc |cc}
\toprule
\multirow{2}{*}{\textbf{Dataset}} & \textbf{Parametric} & \multicolumn{3}{c|}{\textbf{Force Supervision}} & \multicolumn{2}{c}{\textbf{Scale}} \\
\cmidrule(lr){3-5} \cmidrule(lr){6-7}
& \textbf{Body Model} & GRF source & Torque & Uncertainty of Torque & Subjects & Hours \\
\midrule
UnderPressure~\cite{mourot2022underpressure}      & \xmark          & Insole$^*$       & \xmark & \xmark & 10  & 5.5  \\
VidSole~\cite{kambhamettu2025vidsole}             & \xmark        & Insole+Force plate       & \cmark\chg{$^\dagger$} & \xmark & 52  & 6.6   \\
GroundLink~\cite{han2023groundlink}               & SMPL-X        & Force plate  & \xmark & \xmark & 7   & 1.5  \\
ImDy~\cite{liu2025imdy}                           & SMPL          & Simulated    & \cmark & \xmark & 435  & 152.3  \\
AddBiomechanics~\cite{werling2024addbiomechanics} & \xmark        & Force plate  & \cmark & \xmark & 273 & 57.6 \\
\midrule
\textbf{ForceBody (ours)}                          & SKEL & Force plate & \cmark & \cmark & 140 & \chg{26.9} \\
\bottomrule
\end{tabular}
\end{table}


This paper presents \textbf{\textbf{ForceBody}} (see Fig.~\ref{fig:teaser}), a large-scale dataset that \red{aligns} biomechanical force measurements \red{from AddBiomechanics~\cite{werling2023addbiomechanics}} with a parametric body model, and provides uncertainty annotations for every torque label. Our contributions are:

\begin{itemize}
\item \textbf{A force-paired parametric body dataset.} ForceBody comprises \red{10,386} motion trials (9.7M frames, 26.9 hours) from 140 subjects, with per-trial SKEL poses, per-subject body shape parameters, and associated measured GRF and joint torques. To our knowledge, this is the first dataset to pair large-scale \red{measured} forces with a parametric body model. 

\item \textbf{Per-sample uncertainty annotations.} We provide per-frame, per-joint uncertainty annotations for every joint torque label, derived from Monte Carlo sampling through inverse dynamics.  This makes ForceBody the first force dataset to ship with annotated uncertainty.


\item \textbf{Reference baselines for force prediction.} We evaluate six architectures (MLP, Conv1D, LSTM, GRU, Mamba2, Transformer) on the tasks of ground reaction force and joint torque prediction from SKEL motion. As one example use of our uncertainty annotation, finetuning a Transformer with per-sample uncertainty weighting reduces joint torque MAE by $5\%$.

\end{itemize}

\begin{figure}[!t]
\centering
\includegraphics[width=\textwidth]{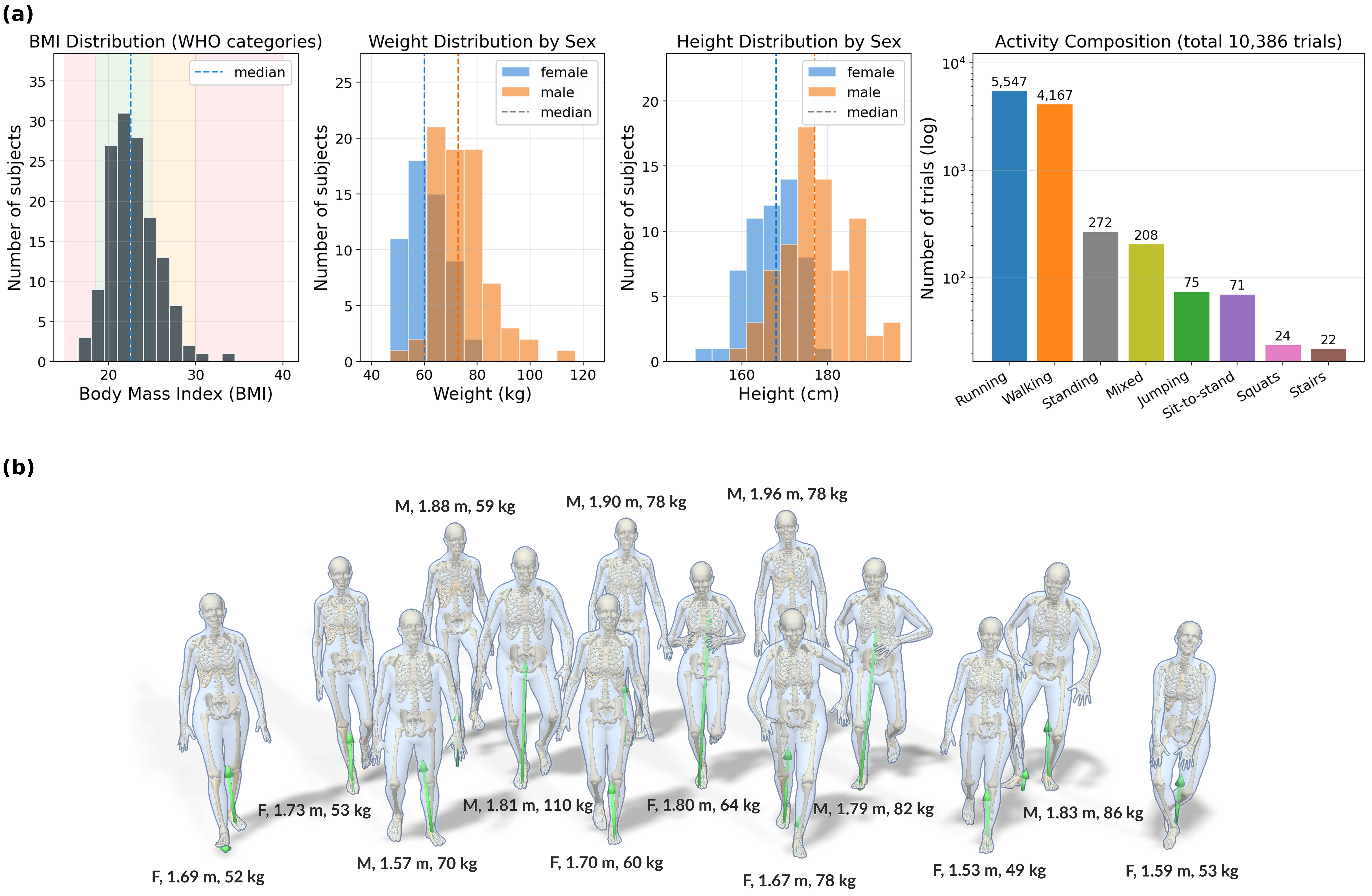}
\caption{\textbf{Dataset statistics of ForceBody.}
(a) Subject anthropometrics (BMI, weight, and height by sex) across 140 subjects
($55$ female, $75$ male, $10$ unreported) and activity composition. In the BMI panel,
pink marks underweight or obese, green marks normal, and orange marks overweight.
The dashed line is the cohort median.
(b) Fourteen ForceBody subjects rendered in one scene at true metric scale with their
own SKEL body shapes, skeletons, and measured ground reaction forces (green arrows).
Labels give sex, height, and body mass.}
\label{fig:dataset_stats}
\end{figure}


\section{Related Work}

\paragraph{Datasets with Force Annotation} Motion datasets vary widely in the type and scale of physical supervision they provide. We compare existing datasets along three factors: the parametric body model, the availability of force supervision, and the scale (Tab.~\ref{tab:dataset_comparison}). These factors are rarely satisfied simultaneously. AddBiomechanics~\cite{werling2024addbiomechanics} offers force-plate-measured GRF and inverse-dynamics torques at scale, but its motion is released as marker trajectories rather than parametric body parameters. ImDy~\cite{liu2025imdy} reaches scale in a parametric representation but its forces come from physics simulation. At smaller scale, UnderPressure~\cite{mourot2022underpressure} and VidSole~\cite{kambhamettu2025vidsole} do not use a parametric body model, while GroundLink~\cite{han2023groundlink} adopts SMPL-X~\cite{pavlakos2019expressive} but does not provide joint torques. None of these datasets, however, measures the reliability of its force annotations.

Beyond force-paired data, parallel lines of work pair parametric motion with biomechanical skeletal anatomy or muscle activations. BioAMASS~\cite{keller2023skin} fits SKEL~\cite{keller2023skin}, a biomechanically grounded body model, to AMASS sequences, but provides no force supervision. Compared to SMPL, SKEL replaces non-anatomical joints with biomechanically grounded ones. 
Since per-joint torques are meaningful only on anatomically grounded joints,  we adopt SKEL as the parametric body model. Other work pairs parametric motion with muscle activations, either measured via sEMG (Muscles in Action~\cite{chiquier2023muscles}) or simulated via OpenSim (MInT~\cite{schneider2024muscles}). These address a different layer of physical supervision.


\paragraph{Uncertainty Analysis of Inverse Dynamics}
\label{sec:rw_uncertainty}
Inverse-dynamics uncertainty has been studied through Monte Carlo sensitivity analysis~\cite{riemer2008uncertainties, uchida2022conclusion, reinbolt2007patient} that perturbs body segment parameters, joint centers, or marker placements to characterize how kinematic noise propagates into torque estimates. 
These studies establish the methodological foundation but are confined to single-subject, single-trial demonstrations.
AddBiomechanics~\cite{werling2024addbiomechanics} reports aggregate, but not per-sample, pipeline uncertainty. 
We close this gap by performing Monte Carlo uncertainty estimation through the inverse-dynamics pipeline for every trial, producing a per-frame, per-joint label uncertainty $\sigma_\tau$ for each torque label. 


\paragraph{Dynamics in Human Motion Modeling}
Existing work has leveraged dynamics for improving motion estimation and generation, which falls into two broad categories. Simulation-based methods embed human models in a physics simulator and apply learned control to ensure the motion complies with physical laws~\cite{peng2018deepmimic,peng2021amp,yuan2021simpoe,yuan2023physdiff}. Constraint-based methods integrate physics-derived losses or priors to enhance the physical plausibility of kinematic estimation~\cite{shimada2020physcap,Zhang_2024_WACV,zhang2024physpt}. All these methods primarily enforce physics laws without grounding in the underlying biomechanical forces, making it difficult to quantitatively evaluate their physical validity. ForceBody provides this grounding: it can be directly used by these pipelines for both training and validation against measured dynamics.

A separate line of work on modeling dynamics focuses on regressing forces from kinematics under supervised learning, ranging from task-specific gait models~\cite{oh2013prediction,mundt2020prediction} to large-scale approaches~\cite{liu2025imdy,liu2025homogeneous,
werling2024addbiomechanics,tan2026gaitdynamics}. These methods primarily estimate forces from sparse marker trajectories~\cite{oh2013prediction,mundt2020prediction,tan2026gaitdynamics}, incompatible with dense parametric body models in vision pipelines. Other methods support SMPL~\cite{han2023groundlink,liu2025imdy} but inherit its biomechanical limitations, small scale, or reliance on simulated forces. ForceBody addresses these limitations by providing large-scale, measured force supervision directly on SKEL.


\begin{figure}[!t]
\centering
\includegraphics[width=\textwidth]{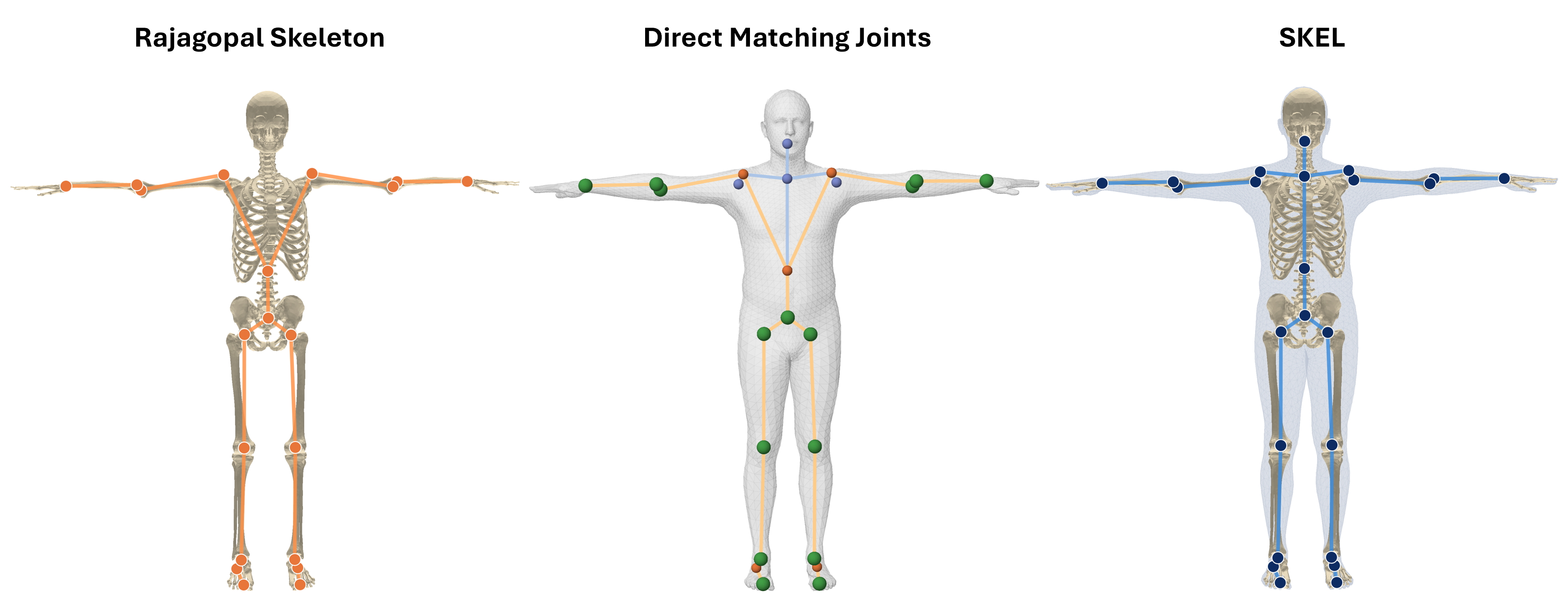}
\caption{\textbf{Joint definitions of AddBiomechanics vs.\ SKEL.}
    \textbf{Left:} Rajagopal skeleton (20 joints, 37 DOFs).
    \textbf{Right:} SKEL (24 joints, 46 DOFs).
    \textbf{Middle:} 15 directly matched joints (green) used as supervision.}
\label{fig:joint_definitions}
\end{figure}

\section{Dataset}

\begin{figure}[!t]
\centering
\includegraphics[width=0.8\textwidth]{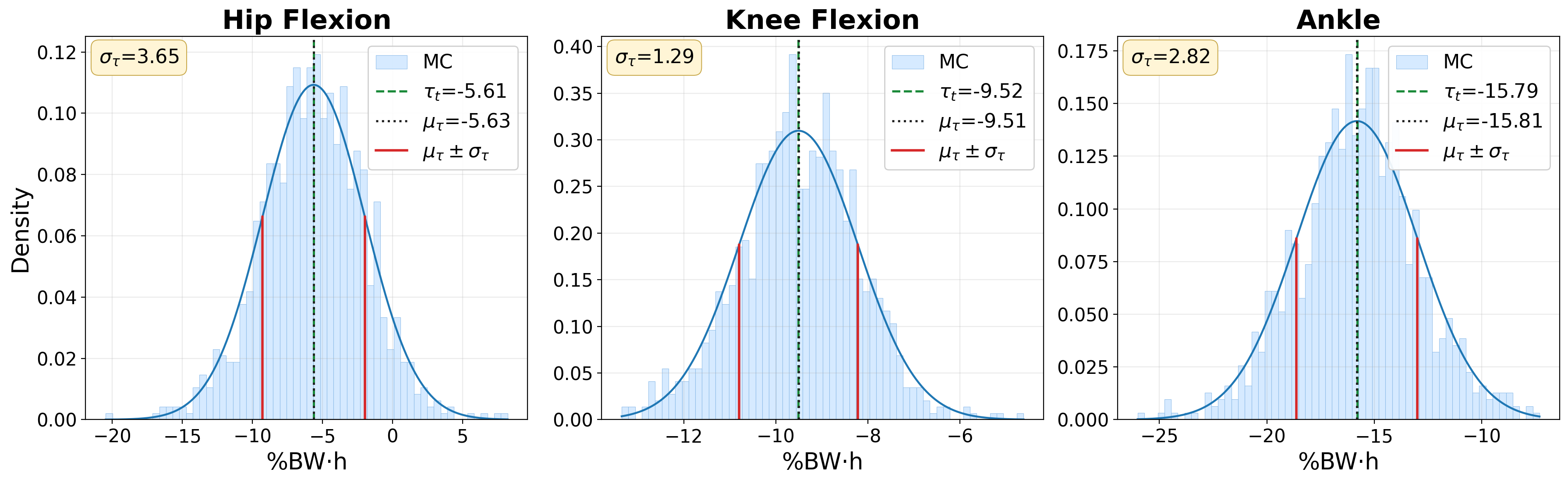}
\caption{\textbf{Per-joint torque uncertainty for one representative frame.} Monte Carlo torque distribution at three right-side joints for a single frame of a Carter2023~\cite{carter2024full} running trial. Each panel shows the histogram of $N\!=\!1{,}000$ MC samples (light blue) with its Gaussian fit. Vertical lines mark the inherited torque $\tau_t$ (green dashed) and the MC mean $\mu_\tau$ (black dotted). The released uncertainty $\sigma_\tau$ is reported in the yellow panel.}
\label{fig:per_joint_sigma_example}
\end{figure}

\begin{table}[!t]
\centering
\caption{\textbf{Released torque uncertainty.} (a) Per-joint median
$\bm{\sigma}_\tau$ across the 14 direct joints with interquartile range (25th to 75th percentile) in brackets. (b) Per-activity median relative ratio $\bm{\sigma}_\tau/|\bm{\tau}|$ over the released \red{$8{,}732$} trials. Each value is the cross-trial median of the per-trial median. A full per-joint, per-activity, and per-source breakdown is provided in Appendix~\ref{app:uncertainty_stats}.}
\label{tab:sigma_combined}

\begin{subtable}[t]{0.48\linewidth}
\centering
\caption{Per joint (14 direct).}
\label{tab:sigma_summary}
\scalebox{0.85}{
\begin{tabular}{l l r}
\toprule
Region & Joint & $\bm{\sigma}_\tau$ (Nm/kg) \\
\midrule
\multirow{3}{*}{Upper body}
 & Elbow      & \red{0.0145\,{\scriptsize[0.0127,\,0.0164]}}     \\
 & Radioulnar & 0.0004\,{\scriptsize[0.0004,\,0.0004]}  \\
 & Wrist      & \red{0.0017\,{\scriptsize[0.0016,\,0.0018]}}   \\
\midrule
\multirow{4}{*}{Lower body}
 & Hip   & \red{0.4658\,{\scriptsize[0.4483,\,0.4936]}}    \\
 & Knee  & \red{0.2108\,{\scriptsize[0.2004,\,0.2242]}}    \\
 & Ankle & \red{0.1156\,{\scriptsize[0.1004,\,0.1252]}}    \\
 & MTP   & \red{0.0009\,{\scriptsize[0.0008,\,0.0010]}}  \\
\bottomrule
\end{tabular}
}
\end{subtable}\hfill
\begin{subtable}[t]{0.48\linewidth}
\centering
\caption{Per activity (\red{8,732} trials).}
\label{tab:sigma_activity}
\renewcommand{\arraystretch}{1.06}
\scalebox{0.85}{
\begin{tabular}{l r r}
\toprule
Activity & Trials & $\bm{\sigma}_\tau/|\bm{\tau}|$ \\
\midrule
\red{Jumping}        & \red{$52$}      & \red{$0.36$} \\
\red{Running}        & \red{$4{,}959$} & \red{$0.50$} \\
\red{Sit-to-stand}   & \red{$29$}      & \red{$0.78$} \\
\red{Others}         & \red{$148$}     & \red{$0.88$} \\
\red{Walking}        & \red{$3{,}310$} & \red{$0.92$} \\
\red{Stairs}         & \red{$4$}       & \red{$1.26$} \\
\red{Standing}       & \red{$230$}     & \red{$1.28$} \\
\bottomrule
\end{tabular}
}
\end{subtable}
\end{table}

\textbf{Dataset statistics.} ForceBody contains \red{10,386} motion trials (9.7M frames at 100\,Hz, 26.9 hours) of 140 subjects (\red{75 male, 55 female}, 10 unreported; height 1.73$\pm$0.09\,m, mass 68.8$\pm$11.5\,kg, BMI 22.8$\pm$2.8\,kg/m$^2$) spanning walking, running, jumping, sit-to-stand, stair climbing, and mixed daily activities (see Fig.~\ref{fig:dataset_stats}). \red{The dataset is dominated by walking and running (93.5\% of trials), with detailed activity and source distribution provided in the Appendix. We adopt a subject-level split stratified by body size, activity, and source, with 121 training subjects (9,472 trials, 24 hours) and 19 disjoint test subjects (914 trials, 3 hours).}

\textbf{Released components.} Each trial in \textbf{ForceBody} contains: (1) SKEL parameters (pose $\bm{\theta}$, shape $\bm{\beta}$, global translation $\mathbf{t}$); (2) measured ground reaction forces (GRFs) \red{with their centers of pressure (CoP), the 2D point on the force plate at which the GRF acts;} (3) inverse-dynamics joint torques $\bm{\tau}$, both inherited from AddB; and (4) per-frame, per-joint uncertainty $\sigma_{\tau}$ on the torques.

To achieve this, we re-express AddBiomechanics' measurement-grounded force data in the SKEL parametric body model and accompany every torque label with a Monte Carlo uncertainty estimated through the same inverse-dynamics pipeline that produced the labels. We first describe the SKEL motion fitting pipeline in Sec.~\ref{sec:fitting}, then the uncertainty annotation of inverse-dynamics torque in Sec.~\ref{sec:uncertainty}. Lastly, we show the fitting results and validity of force and SKEL motion pairing in Sec.~\ref{sec:dataset-analysis}.

\subsection{SKEL Motion Fitting}
\label{sec:fitting}

\textbf{Why SKEL.} We choose SKEL~\cite{keller2023skin} as the parametric body model for \textbf{ForceBody} for two reasons. First, SKEL inherits the 46-DOF biomechanical skeleton (BSM) from the OpenSim community, providing anatomically grounded joint axes that are required for valid inverse-dynamics interpretation of the inherited torque labels. Second, SKEL is fitted inside SMPL bodies, exposing the same surface mesh and shape space as SMPL-based encoders, decoders, and motion priors, which form the dominant system for human motion learning.

\textbf{Problem setting.} Each AddBiomechanics trial provides motion in the OpenSim Rajagopal skeleton~\cite{rajagopal2016full}, encoded as $T$~frames of joint angles together with anatomical joint centers, bone segments, and optical marker trajectories. Our task is to recover a single SKEL shape $\bm{\beta} \in \mathbb{R}^{10}$ \red{per subject} along with per-frame pose $\bm{\theta}_t \in \mathbb{R}^{46}$ and global translation $\mathbf{t}_t \in \mathbb{R}^{3}$ that reproduce the recorded motion under the SKEL kinematic model. Because SKEL is a full-body model, we use only AddB's full-body recordings; the half-body subset lacks the upper-body data needed for fitting.

\textbf{From Rajagopal kinematics to SKEL parameters.} This recovery is underdetermined. AddB exposes only 20 joints (37 DOFs) while SKEL has 24 joints (46 DOFs), so 9 SKEL DOFs at the spine, scapula, and humerus have no direct AddB counterpart and must be recovered through the kinematic chain. Of the 20 AddB joints, 15 have anatomically equivalent counterparts in SKEL (pelvis, bilateral hip, knee, ankle, MTP, elbow, radioulnar, and wrist; Fig.~\ref{fig:joint_definitions}), and serve as direct joint-position supervision. \red{Our fitting recovers $\bm{\beta}$ once per subject and then $\{\bm{\theta}_t, \mathbf{t}_t\}_{t=1}^T$ for each trial.} Our pipeline adapts the SKEL fitting of Keller~{\it et~al.}~\cite{keller2023skin} to AddB, with bone-length and temporal regularizers added to handle the marker and joint layout differences.

\textbf{Stage 1.} We estimate \red{one shape $\bm{\beta} \in \mathbb{R}^{10}$ per subject by matching SKEL bone lengths to the subject's AddB bone lengths, together with terms for the recorded height and body mass. A single shape per subject keeps the body identical across all trials of that person.} $\bm{\beta}$ is then frozen for Stage 2. \textbf{Stage 2.} With $\bm{\beta}$ fixed, we optimize the per-frame pose $\bm{\theta}_t \in \mathbb{R}^{46}$ and translation $\mathbf{t}_t \in \mathbb{R}^{3}$ using two positional supervisions: (i) a joint-position loss on the 15 directly matched joints (Fig.~\ref{fig:joint_definitions}), and (ii) a marker loss that aligns SKEL mesh vertices to the observed optical markers via the SMPL$\to$AddB vertex correspondence~\cite{keller2023skin}. We additionally apply anatomical regularizers on under-constrained joints (spine, scapula, humerus) and a temporal smoothness term on velocity and acceleration. Source recordings span $61$--$250$\,Hz, so we standardize them to $100$\,Hz and use only AddB's IK output (excluding force-based refinement) to prevent force leakage into kinematic inputs~\cite{tan2026gaitdynamics}. The full loss formulation, hyperparameters, and preprocessing details are provided in Appendix~\ref{app:fitting}.


\begin{figure}[!t]
    \includegraphics[width=\linewidth]{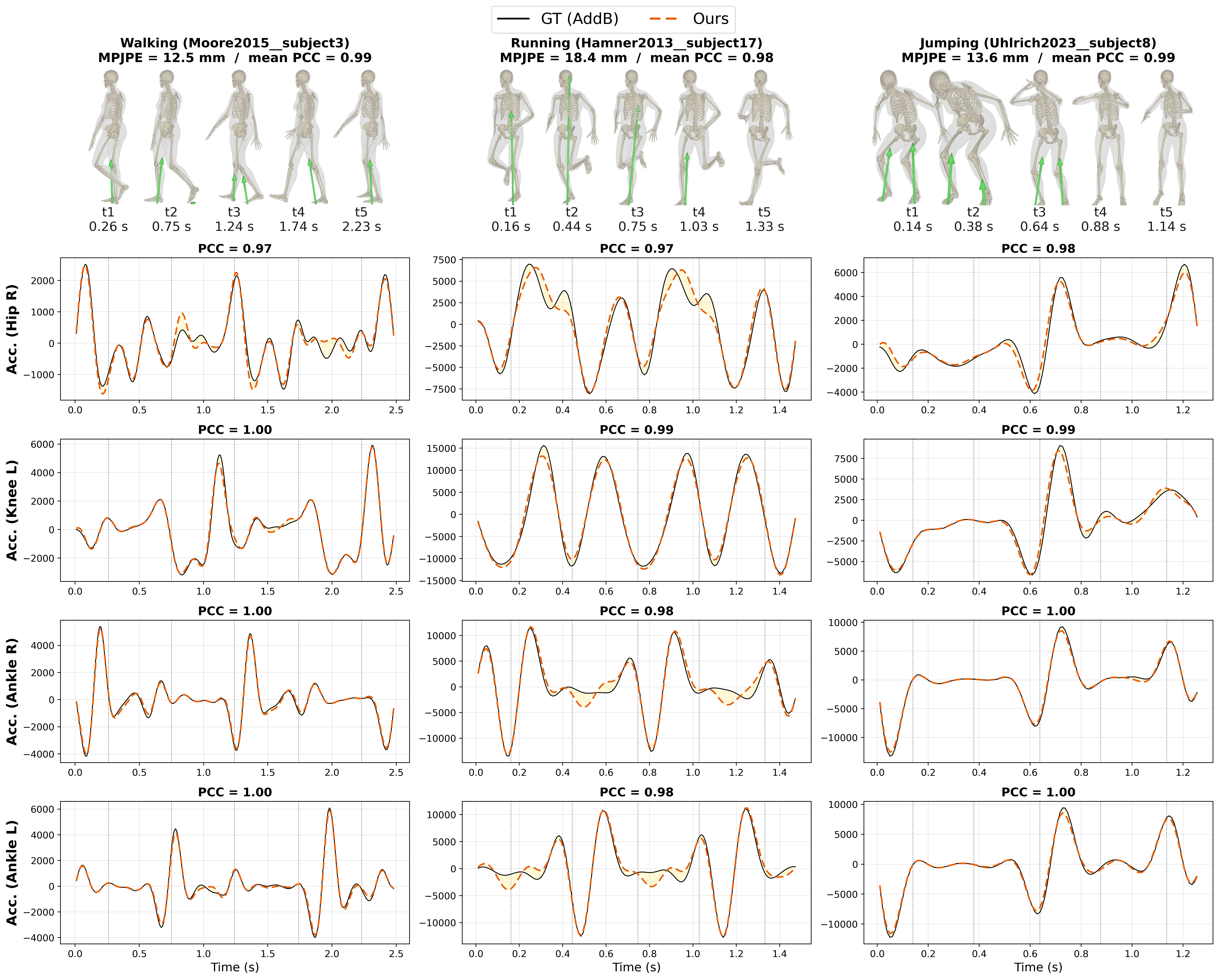}
    \caption{\textbf{DOF acceleration comparison: GT vs.\ ForceBody SKEL fit.} Right hip, left knee, right ankle, and left ankle (top to bottom) angular acceleration (deg/s$^2$) for one trial of each of three activities (Walking, Running, Jumping). PCC reported per panel. The top row shows the fitted SKEL body at five time points t1 to t5, marked by the
dotted lines in the curves below, with measured GRF as green arrows.}
    \label{fig:dof_acc_compare}
\end{figure}

\subsection{Per-Sample Torque Uncertainty}
\label{sec:uncertainty}

For each trial, AddBiomechanics provides joint torques $\bm{\tau}_t$ alongside the captured kinematics. $\bm{\tau}_t$ is not observed but computed by inverse dynamics from joint kinematics $\bm{q}_t$ and the measured GRF $\bm{\lambda}_t$. Inverse dynamics propagates kinematic noise nonlinearly into the output torques~\cite{riemer2008uncertainties,uchida2022conclusion}. We accompany every released $\bm{\tau}_t$ with a per-frame, per-joint uncertainty annotation $\sigma_{\tau,t}$.

We obtain $\sigma_\tau$ by Monte Carlo sampling through the same OpenSim inverse-dynamics solver that produced the inherited torques. At each frame we perturb the AddB joint angles with band-limited Gaussian noise $\bm{\epsilon}\!\sim\!\mathcal{N}(\bm{0},\sigma_q^2\mathbf{I})$ at the 20 direct DOFs spanning the 14 major joints (bilateral hip, knee, ankle, MTP, elbow, radioulnar, wrist), feed each perturbed pose back through the solver with GRF $\bm{\lambda}_t$ held fixed, and report the per-DOF sample mean $\mu_\tau$ and standard deviation $\sigma_\tau$ over $N$ samples. The full algorithm is in Appendix~\ref{app:mc_algo}. We set $\sigma_q\!=\!1.6^{\circ}$, matching the average joint-angle RMSE that AddB reports for its inverse-kinematics pass on synthetic walking data~\cite{werling2023addbiomechanics}, so that the perturbations match the noise level the released kinematics actually carry. Because AddB reports only an aggregate angle error and does not release per-DOF noise levels, we apply $\sigma_q$ uniformly to all 20 direct DOFs as an isotropic approximation. We use $N\!=\!1{,}000$ samples per frame. The standard error of $\hat{\sigma}_\tau$ shrinks from $7.1\%$ at $N\!=\!100$ to $3.2\%$ at $N\!=\!500$ and $2.2\%$ at $N\!=\!1{,}000$, beyond which further increases offer marginal returns. Of the \red{$10{,}386$} trials, we exclude \red{$1{,}654$} (\red{$15.9\%$}) with missing GRF labels or coverage below $50\%$, leaving \red{$8{,}732$} trials with released $\sigma_\tau$.

Fig.~\ref{fig:per_joint_sigma_example} visualizes the per-joint MC histograms for one representative frame from a Carter2023~\cite{carter2024full} running trial. As an aggregate-level sanity check, the median released $\bm{\sigma}_\tau$ on walking trials is approximately $0.10\,\%$BW$\cdot$h, on the same order as the $0.15\,\%$BW$\cdot$h walking-torque error AddBiomechanics reports for its own pipeline~\cite{werling2023addbiomechanics}. Across the 14 direct joints, the released $\sigma_\tau$ spans about \red{three} orders of magnitude (Tab.~\ref{tab:sigma_summary}). The distal upper-body joints (radioulnar, wrist) sit at the bottom of the range, since they only swing light distal segments and the same angle perturbation changes very little torque. The elbow sits in the middle of the range. Lower-body proximal joints (hip, knee) sit at the top of the range, since they carry the entire body weight
through the longest lever arms, so a small angle perturbation moves a large torque. The toe joint (MTP) is the lower-body exception, with a $\sigma_\tau$ comparable to the upper-body distal joints. More detailed analyses on per-activity and per-source breakdowns are in Appendix~\ref{app:uncertainty_stats}.

Beyond joint-level patterns, the relative ratio $\bm{\sigma}_\tau/|\bm{\tau}|$ is motion dependent (Tab.~\ref{tab:sigma_activity}). Dynamic motions yield smaller relative uncertainty (around \red{$0.4$ in jumping and $0.5$ in running}) than near-static ones (around $1.3$ in \red{standing}), with \red{sit-to-stand and walking} falling in between. Across all trials, $\sigma_\tau/|\tau|$ is strongly anti-correlated with both joint-velocity RMS (Spearman $\rho\!=\!$\red{$-0.85$}) and joint-torque RMS (Spearman $\rho\!=\!$\red{$-0.81$}). The high ratio in static activities reflects the small underlying $|\bm{\tau}|$ ($\sim\!$\red{$0.017$}\,Nm/kg) rather than inflated $\bm{\sigma}_\tau$.


\begin{figure*}[!t]
\centering
\includegraphics[width=\dimexpr\textwidth-3pt\relax]{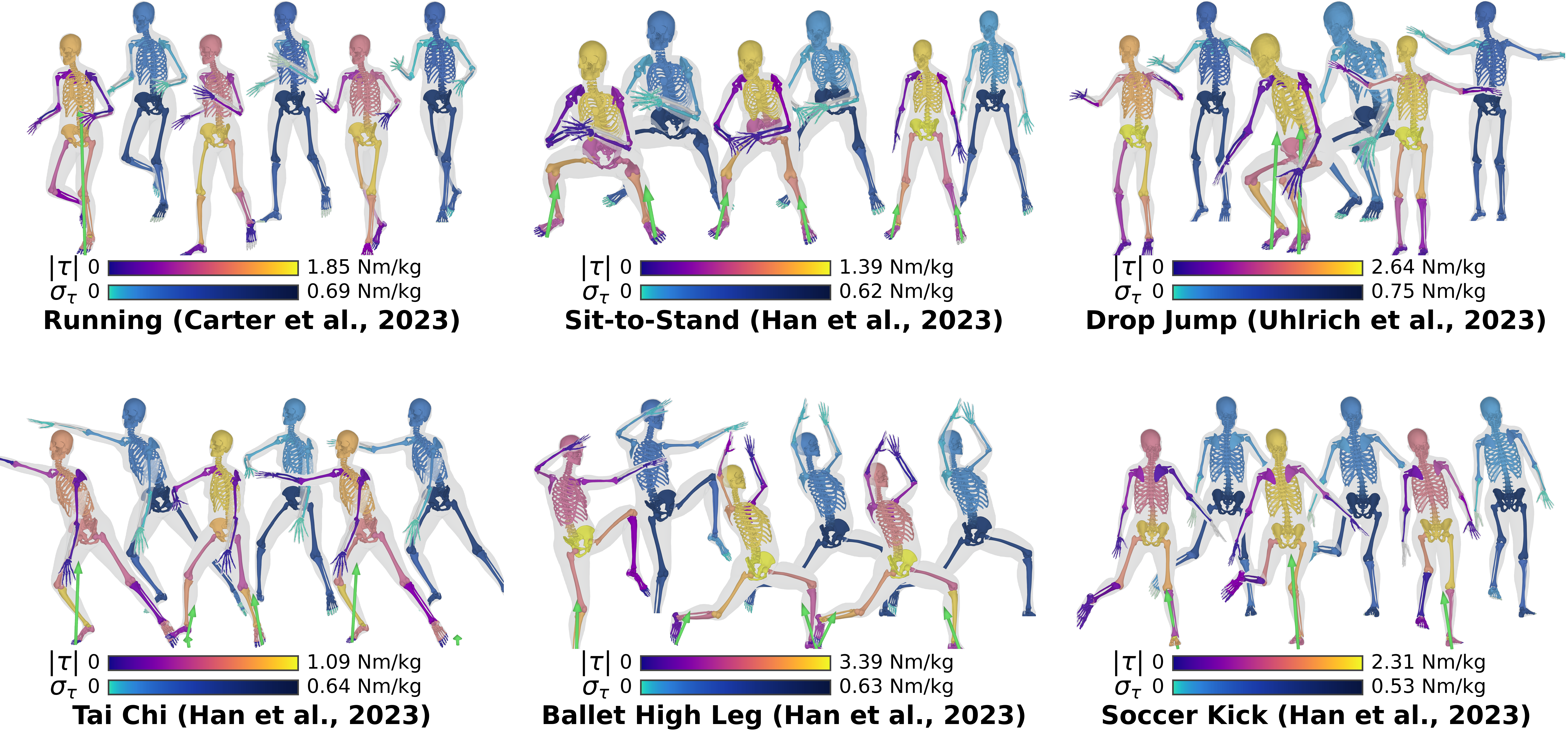}
\caption{\textbf{Torques and per-frame torque uncertainty across activities.} Each panel shows three
frames of one activity (Running, Sit-to-Stand, Drop Jump, Tai Chi, Ballet High Leg, Soccer Kick). Foreground skeleton is colored by $|\tau|$ and the background skeleton is colored by
$\sigma_\tau$ from MC samples. Colorbars below each panel: $|\tau|$ (top), $\sigma_\tau$ (bottom), both in Nm/kg. Green arrows are measured GRF.}
\label{fig:qualitative}
\end{figure*}

\subsection{Pairing Torques with SKEL Motions}
\label{sec:dataset-analysis}

ForceBody is not just a fitting dataset; its value rests on the pairing between fitted SKEL motion and inherited torque. We assess the validity of our pairing along two axes: motion fidelity with AddB (Motion Fitting Quality) and consistency between SKEL-derived and AddB torques (Torque Consistency).

\textbf{Motion Fitting Quality.} The released SKEL fits track the AddB kinematics closely. Mean per-joint Euclidean error is \red{$21.6$}\,mm overall. We also report standard physical plausibility metrics: foot sliding (\red{$1.23$}\,mm) and ground penetration (\red{$1.17$}\,mm). For acceleration, Fig.~\ref{fig:dof_acc_compare} compares finite-difference angular acceleration between AddB and our SKEL fits on walking, running, and jumping trials. The two curves stay on top of each other across all panels with Pearson correlation between \red{$0.97$} and $1.00$, so the fit captures the underlying acceleration profile rather than only matching frame-level joint positions. Fig.~\ref{fig:qualitative} shows SKEL meshes, internal skeletons, joint-torque heatmaps, and measured GRF arrows for one trial of each major activity (running, sit-to-stand, drop jump, tai chi, ballet high leg, and soccer kick), illustrating the released motions across range of behaviors covered. Per-joint and physical plausibility statistics are tabulated in Appendix~\ref{app:fitting_quality}. 

\begin{figure}[!t]
    \centering
    \includegraphics[width=\linewidth]{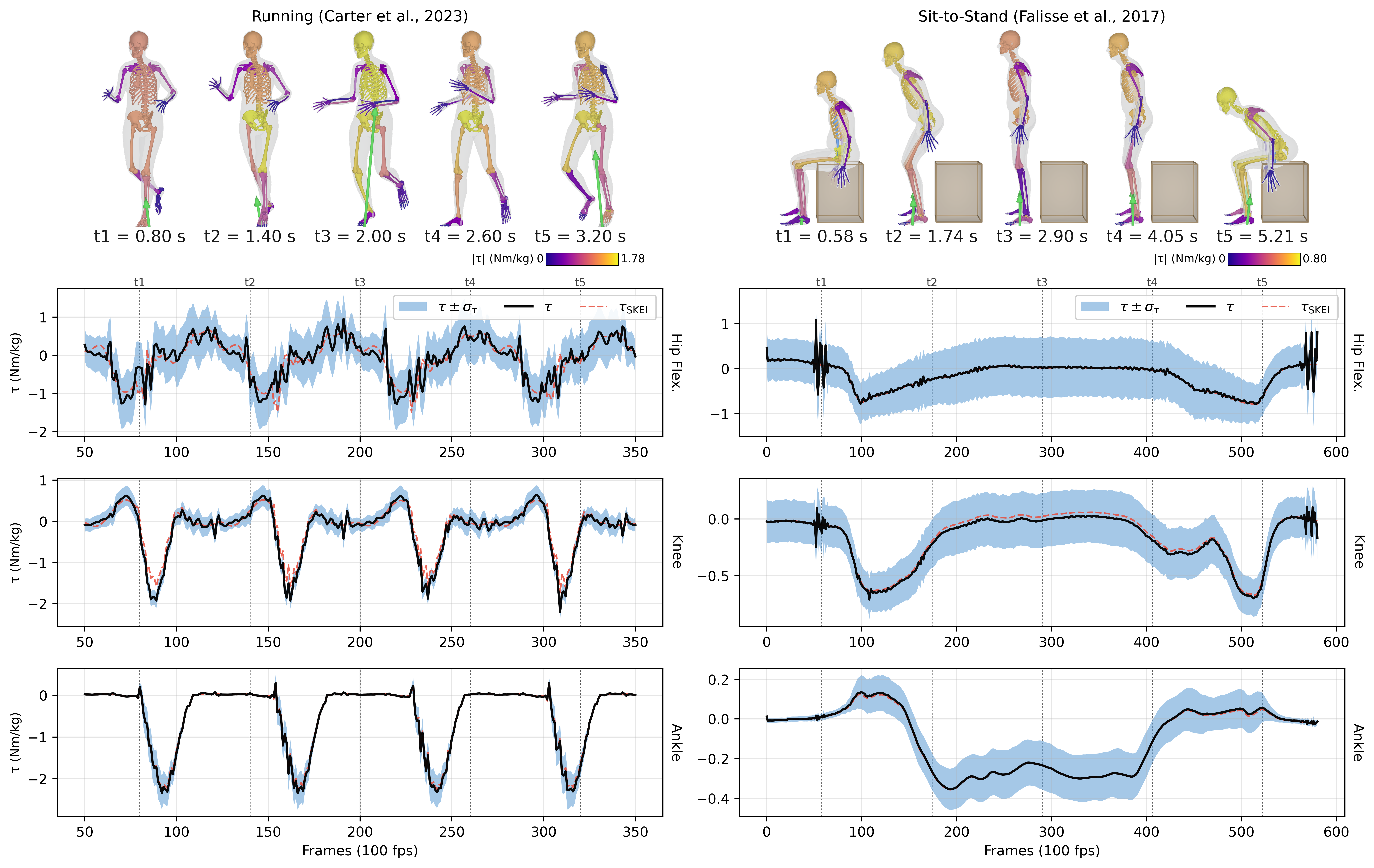}
    \caption{\textbf{Per-joint torque trajectories with MC uncertainty.}
    Hip, knee, and ankle torques on a representative running (left;
    Carter2023~\cite{carter2024full}) and sit-to-stand (right;
    Falisse2017~\cite{falisse2016emg}) trial. $\tau_{\mathrm{SKEL}}$ (red
    dashed) tracks $\tau$ (black solid) inside the MC band $\tau \pm
    \sigma_\tau$ (shaded blue) on the plotted joints. The top row shows the SKEL body at five time points t1 to t5 (dotted lines), with bones colored by $|\tau|$ (Nm/kg, colorbar below the row) and measured GRF as green
arrows.}
    \label{fig:uncertainty}
\end{figure}

\textbf{Torque Consistency.}
To verify the SKEL fit at the torque level, we feed the fitted SKEL kinematics through the same OpenSim inverse-dynamics pipeline that produced $\tau$, obtaining a SKEL-derived torque $\tau_\text{SKEL}$.  Fig.~\ref{fig:uncertainty} compares $\tau_\text{SKEL}$ with the inherited $\tau$ against the $\tau \pm \sigma_\tau$ band on representative running and sit-to-stand trials; $\tau_\text{SKEL}$ tracks $\tau$ closely, indicating that the SKEL fit recovers torque-level dynamics consistent with the inherited labels. \red{Across the trials with released $\sigma_\tau$ ($8{,}455$ of $8{,}732$ after an inverse-kinematics residual check), the median per-trial RMSE between $\tau_\text{SKEL}$ and $\tau$ is $0.31$, $0.20$, and $0.04$\,Nm/kg at the hip, knee, and ankle. $\tau_\text{SKEL}$ falls inside the $\tau \pm \sigma_\tau$ band on $89\%$, $83\%$, and $89\%$ of the frames at these joints. The gap is the largest in running (hip RMSE $0.38$\,Nm/kg) and smallest in standing (hip RMSE $0.03$\,Nm/kg, $99\%$ inside the band at hip and knee). Upper-body joints show lower coverage ($41$ to $64\%$), where the SKEL and Rajagopal joint axes differ the most (Fig.~\ref{fig:joint_definitions}).}


\section{Experiments}
\label{sec:experiments}

We use ForceBody to study two questions. Firstly, \textit{does ForceBody's parametric representation match the predictive utility of AddBiomechanics' Rajagopal-based pipelines?} (Sec.~\ref{sec:compare addb}) Secondly, \textit{what level of torque prediction performance can be achieved on ForceBody, and how can the released uncertainty annotation be used?} (Sec.~\ref{sec:baselines}).


\subsection{Comparison with AddB}
\label{sec:compare addb}
Following AddBiomechanics' baseline protocol~\cite{werling2024addbiomechanics}, we predict per-foot ground reaction force (GRF) and per-foot ground reaction moment (GRM) from SKEL poses $\bm{\theta}$. For comparison, we train MLP and GLN, the architectures used by AddB~\cite{werling2024addbiomechanics}, on ForceBody. Both pipelines take as input the same underlying motion, expressed in different parametric representations: The AddB baselines use Rajagopal joint kinematics from AddBiomechanics, while our models use SKEL joint kinematics fitted on the same trials. As shown in Tab.~\ref{tab:contact_dynamics}, across matched architectures, ForceBody-trained models perform on par with their AddB-trained counterparts. The SKEL parametric representation thus supports force prediction at AddB-level performance.

\subsection{Reference Baselines for Torque Prediction}
\label{sec:baselines}

\paragraph{Joint Torque Prediction}

Given a sequence of SKEL pose parameters $\bm{q}_{1:T}$ and shape parameters $\bm{\beta}$, the model predicts the AddB inverse-dynamics torque $\bm{\tau}_{1:T}$ at every frame, mass-normalized to Nm/kg. We evaluate on the \red{14} anatomical joints \red{(20 DOFs)} with a direct SKEL correspondence, and compare six architectures (MLP, Conv1D, LSTM, GRU, Mamba2, Transformer) trained on ForceBody. We report errors separately on the upper body, the lower body, and as an overall \textit{All joints} mean, using RMSE, MAE, and Pearson correlation. Each baseline is fully supervised on the released torque annotations. All six backbones produce plausible torque predictions (Tab.~\ref{tab:tau_aggregate}), and Mamba2 performs the best. Adding the parametric body inputs (SKEL $\bm{\beta}$, body mass, and height) \red{to} \red{the} pose \red{does} \red{not} \red{reduce} \red{the error, while adding pose} and \red{root} \red{velocities} \red{and} \red{accelerations raises the \textit{All joints} PCC from $0.67$ to $0.76$} (Appendix~\ref{app:input_ablation}). Detailed training settings and the input feature ablation are in Appendix~\ref{app:training} and~\ref{app:input_ablation}.

\begin{table}[t!]
\centering
\setlength{\abovecaptionskip}{0pt}
\setlength{\belowcaptionskip}{8pt}
\caption{\textbf{ForceBody matches AddB on per-foot GRF and GRM.} GRF: ground reaction force; GRM: ground reaction moment. Architecture-matched comparison. All errors are mass-normalized.}
\label{tab:contact_dynamics}
\small
\renewcommand{\arraystretch}{1.0}
\setlength{\tabcolsep}{6pt}
\scalebox{0.85}{
\begin{tabular}{lllcc}
\toprule
\textbf{Dataset} & \textbf{Body Model} & \textbf{Architecture} & \textbf{GRF} (N/kg, $\downarrow$) & \textbf{GRM} (Nm/kg, $\downarrow$) \\
\midrule
\multirow{2}{*}{AddB} & \multirow{2}{*}{Rajagopal~\cite{rajagopal2016full}}
 & MLP & 2.091 & 0.278 \\
 & & GLN & 1.712 & 0.275 \\
\midrule
\multirow{2}{*}{ForceBody} & \multirow{2}{*}{SKEL~\cite{keller2023skin}}
   & MLP & \red{1.165\,{\scriptsize$\pm$0.011}} & \red{0.168\,{\scriptsize$\pm$0.009}} \\
  & & GLN & \red{1.136\,{\scriptsize$\pm$0.040}} & \red{0.149\,{\scriptsize$\pm$0.011}} \\
\bottomrule
\end{tabular}
}
\end{table}

\begin{table}[t]
\centering
\setlength{\abovecaptionskip}{0pt}
\setlength{\belowcaptionskip}{8pt}
\caption{\textbf{Joint torque $\tau$ prediction} on ForceBody. \textit{Upper mean} = elbow, radioulnar, wrist (R/L averaged). \textit{Lower mean} = hip, knee, ankle, MTP (R/L averaged). Values are mean $\pm$ standard deviation over three seeds.}
\label{tab:tau_aggregate}
\small
\setlength{\tabcolsep}{2pt}
\renewcommand{\arraystretch}{1.05}
\resizebox{\linewidth}{!}{\begin{tabular}{l|ccc|ccc|ccc}
\toprule
\multirow{2}{*}{} & \multicolumn{3}{c|}{\textbf{MLP}} & \multicolumn{3}{c|}{\textbf{Conv1D}} & \multicolumn{3}{c}{\textbf{LSTM}~\cite{hochreiter1997long}} \\
\cmidrule(lr){2-4} \cmidrule(lr){5-7} \cmidrule(lr){8-10}
 & R$\downarrow$ & M$\downarrow$ & P$\uparrow$ & R$\downarrow$ & M$\downarrow$ & P$\uparrow$ & R$\downarrow$ & M$\downarrow$ & P$\uparrow$ \\
\midrule
\textbf{Upper mean} & \red{0.008\,{\scriptsize$\pm$0.000}} & \red{0.004\,{\scriptsize$\pm$0.000}} & \red{0.644\,{\scriptsize$\pm$0.002}} & \red{0.007\,{\scriptsize$\pm$0.000}} & \red{0.004\,{\scriptsize$\pm$0.000}} & \red{0.700\,{\scriptsize$\pm$0.003}} & \red{0.007\,{\scriptsize$\pm$0.000}} & \red{0.004\,{\scriptsize$\pm$0.000}} & \red{0.691\,{\scriptsize$\pm$0.004}} \\
\textbf{Lower mean} & \red{0.241\,{\scriptsize$\pm$0.006}} & \red{0.168\,{\scriptsize$\pm$0.004}} & \red{0.833\,{\scriptsize$\pm$0.003}} & \red{0.210\,{\scriptsize$\pm$0.003}} & \red{0.141\,{\scriptsize$\pm$0.002}} & \red{0.870\,{\scriptsize$\pm$0.003}} & \red{0.214\,{\scriptsize$\pm$0.004}} & \red{0.145\,{\scriptsize$\pm$0.002}} & \red{0.864\,{\scriptsize$\pm$0.003}} \\
\midrule
\textbf{All joints}  & \red{0.148\,{\scriptsize$\pm$0.003}} & \red{0.103\,{\scriptsize$\pm$0.002}} & \red{0.757\,{\scriptsize$\pm$0.002}} & \red{0.129\,{\scriptsize$\pm$0.002}} & \red{0.086\,{\scriptsize$\pm$0.001}} & \red{0.802\,{\scriptsize$\pm$0.002}} & \red{0.131\,{\scriptsize$\pm$0.002}} & \red{0.088\,{\scriptsize$\pm$0.001}} & \red{0.795\,{\scriptsize$\pm$0.003}} \\
\bottomrule
\end{tabular}}
\\[6pt]
\resizebox{\linewidth}{!}{\begin{tabular}{l|ccc|ccc|ccc}
\toprule
\multirow{2}{*}{} & \multicolumn{3}{c|}{\textbf{GRU}~\cite{cho2014learning}} & \multicolumn{3}{c|}{\textbf{Mamba2}~\cite{dao2024transformers}} & \multicolumn{3}{c}{\textbf{Transformer}~\cite{vaswani2017attention}} \\
\cmidrule(lr){2-4} \cmidrule(lr){5-7} \cmidrule(lr){8-10}
 & R$\downarrow$ & M$\downarrow$ & P$\uparrow$ & R$\downarrow$ & M$\downarrow$ & P$\uparrow$ & R$\downarrow$ & M$\downarrow$ & P$\uparrow$ \\
\midrule
\textbf{Upper mean} & \red{0.007\,{\scriptsize$\pm$0.000}} & \red{0.003\,{\scriptsize$\pm$0.000}} & \red{0.701\,{\scriptsize$\pm$0.009}} & \red{0.007\,{\scriptsize$\pm$0.000}} & \red{0.003\,{\scriptsize$\pm$0.000}} & \red{0.714\,{\scriptsize$\pm$0.005}} & \red{0.008\,{\scriptsize$\pm$0.000}} & \red{0.005\,{\scriptsize$\pm$0.000}} & \red{0.651\,{\scriptsize$\pm$0.014}} \\
\textbf{Lower mean} & \red{0.207\,{\scriptsize$\pm$0.002}} & \red{0.139\,{\scriptsize$\pm$0.002}} & \red{0.875\,{\scriptsize$\pm$0.002}} & \red{0.204\,{\scriptsize$\pm$0.002}} & \red{0.134\,{\scriptsize$\pm$0.002}} & \red{0.878\,{\scriptsize$\pm$0.004}} & \red{0.253\,{\scriptsize$\pm$0.015}} & \red{0.176\,{\scriptsize$\pm$0.011}} & \red{0.833\,{\scriptsize$\pm$0.007}} \\
\midrule
\textbf{All joints}  & \red{0.127\,{\scriptsize$\pm$0.001}} & \red{0.085\,{\scriptsize$\pm$0.001}} & \red{0.805\,{\scriptsize$\pm$0.003}} & \red{0.125\,{\scriptsize$\pm$0.001}} & \red{0.082\,{\scriptsize$\pm$0.001}} & \red{0.813\,{\scriptsize$\pm$0.004}} & \red{0.155\,{\scriptsize$\pm$0.009}} & \red{0.107\,{\scriptsize$\pm$0.007}} & \red{0.760\,{\scriptsize$\pm$0.010}} \\
\bottomrule
\end{tabular}}
\\[2pt]
\begin{flushleft}\footnotesize R: RMSE (Nm/kg) \quad M: MAE (Nm/kg) \quad P: PCC.\end{flushleft}
\end{table}

\paragraph{Torque Prediction with Uncertainty-guidance}
To demonstrate one use of the released per-frame, per-joint uncertainty annotations, we enhance the torque prediction model with an uncertainty-aware loss that \red{assigns each training window a loss weight computed from its mean $\sigma_\tau$}~\cite{zhang2023body}. Following recent force and dynamics prediction models~\cite{tan2026gaitdynamics,schneider2024muscles}, we adopt a Transformer backbone. Compared to its performance without uncertainty (Tab.~\ref{tab:tau_aggregate_p80_uw}), we improve the \textit{All joints} RMSE from \red{$0.155$ to $0.149$} Nm/kg \red{($-4\%$)}, MAE from \red{$0.107$ to $0.101$ Nm/kg ($-5\%$)}, and PCC from \red{$0.760$ to $0.772$ ($+2\%$), averaged over three seeds,} with gains on both upper- and lower-body joint groups. Because the only change is the per-sample loss weight, the gain comes from the $\sigma_\tau$ annotation alone. These results lay a foundation for further uncertainty-aware methods that the community can build on top of ForceBody.

\begin{table*}[t!]
\centering
\setlength{\abovecaptionskip}{0pt}
\setlength{\belowcaptionskip}{8pt}
\caption{\textbf{Joint torque $\tau$ prediction} on ForceBody. \textit{Upper mean} = elbow/radioulnar/wrist. \textit{Lower mean} = hip/knee/ankle/mtp. \textbf{With Uncertainty} columns: the same Transformer architecture finetuned with torque uncertainty. Mean over three seeds.}
\label{tab:tau_aggregate_p80_uw}
\setlength{\tabcolsep}{3pt}
\scalebox{0.8}{
\begin{tabular}{l|ccc||ccc}
\toprule
\multirow{2}{*}{\textbf{}} & \multicolumn{3}{c||}{\textbf{Without Uncertainty}} & \multicolumn{3}{c}{\textbf{With Uncertainty}} \\
\cmidrule(lr){2-4} \cmidrule(lr){5-7}
 & R$\downarrow$ & M$\downarrow$ & P$\uparrow$ & R$\downarrow$ & M$\downarrow$ & P$\uparrow$ \\
\midrule
\textbf{Upper mean} & \red{0.008\,{\scriptsize$\pm$0.000}} & \red{0.005\,{\scriptsize$\pm$0.000}} & \red{0.651\,{\scriptsize$\pm$0.014}} & \red{0.008\,{\scriptsize$\pm$0.000}\,\textcolor{brightpurple}{($-2\%$)}} & \red{0.004\,{\scriptsize$\pm$0.000}\,\textcolor{brightpurple}{($-4\%$)}} & \red{0.666\,{\scriptsize$\pm$0.014}\,\textcolor{brightpurple}{($+2\%$)}} \\
\textbf{Lower mean} & \red{0.253\,{\scriptsize$\pm$0.015}} & \red{0.176\,{\scriptsize$\pm$0.011}} & \red{0.833\,{\scriptsize$\pm$0.007}} & \red{0.243\,{\scriptsize$\pm$0.012}\,\textcolor{brightpurple}{($-4\%$)}} & \red{0.166\,{\scriptsize$\pm$0.011}\,\textcolor{brightpurple}{($-5\%$)}} & \red{0.842\,{\scriptsize$\pm$0.011}\,\textcolor{brightpurple}{($+1\%$)}} \\
\midrule
\textbf{All joints} & \red{0.155\,{\scriptsize$\pm$0.009}} & \red{0.107\,{\scriptsize$\pm$0.007}} & \red{0.760\,{\scriptsize$\pm$0.010}} & \red{0.149\,{\scriptsize$\pm$0.007}\,\textcolor{brightpurple}{($-4\%$)}} & \red{0.101\,{\scriptsize$\pm$0.007}\,\textcolor{brightpurple}{($-5\%$)}} & \red{0.772\,{\scriptsize$\pm$0.012}\,\textcolor{brightpurple}{($+2\%$)}} \\
\bottomrule
\end{tabular}
}
\\[2pt]
\begin{flushleft}\footnotesize R: RMSE (Nm/kg) \quad M: MAE (Nm/kg) \quad P: PCC.\end{flushleft}
\end{table*}


\section{Conclusion}
\label{sec:conclusion}

We presented \textbf{ForceBody}, the first large-scale dataset to pair measured ground reaction force (GRF) and inverse-dynamics torques with the SKEL parametric body, and the first to release per-frame, per-joint Monte Carlo uncertainty for every torque label $\sigma_\tau$. Benchmarking \red{GRF prediction against the AddBiomechanics baselines and joint-torque prediction across six architectures} shows that ForceBody enables direct, end-to-end learning of physical signals on a parametric body. As one example use of the released $\sigma_\tau$, fine-tuning a Transformer with per-sample uncertainty weighting yields gains across joint groups. Together, ForceBody opens a path toward force-grounded, uncertainty-aware modeling of human dynamics.

One limitation of our work lies in the representational capacity. The current fitting stage uses SKEL as the parametric body model. Atypical body shapes outside the SKEL shape space can be addressed by extending our pipeline to new parametric models with different shape spaces. \red{The released $\sigma_\tau$ models joint-angle noise only and does not cover GRF and CoP errors, segment inertia, or the axis mismatch between SKEL and the Rajagopal model. The data are dominated by walking and running (Appendix~\ref{app:dataset_composition}), so results on rare activities rest on few trials.}



\begin{ack}
This project was supported by Michigan State University.
\end{ack}


\bibliographystyle{splncs04}
\bibliography{main}


\clearpage
\appendix
\newpage
\section*{Appendix Contents}
\begin{itemize}
  \item \ref{app:dataset_composition}~Per-Dataset Composition
  \item \ref{app:fitting}~Fitting Pipeline
  \item \ref{app:mc_algo}~Pseudo-code for Per-frame Torque Uncertainty
  \item \ref{app:uncertainty_stats}~Per-Sample Torque Uncertainty Statistics
  \item \ref{app:filtering}~Quality-based Filtering
  \item \ref{app:fitting_quality}~Fitting Quality of ForceBody
  \item \ref{app:training}~Joint Torque Training Details
  \item \ref{app:perjoint}~Per-Joint Torque Prediction
  \item \ref{app:input_ablation}~Input Feature Ablation
\end{itemize}
\vspace{1em}


\section{Per-Dataset Composition}

\label{app:dataset_composition}

\red{\textbf{Composition and imbalance.} ForceBody inherits a strong imbalance from its sources. Carter2023~\cite{carter2024full} contributes 7,006 of the 10,386 trials (67.5\%), and running (5,547 trials) and walking (4,167 trials) together make up 93.5\% of the data. The other classes are small. Standing, including quiet standing, one-leg standing, and stance transitions, has 272 trials, jumping 75, sit-to-stand 71, squats 24, and stairs 22. Mixed activities (yoga, gait transitions, sport, dance, tai chi, and stretching) add 208 trials. The cohort has 75 male and 55 female subjects. The 10 subjects of vanderZee2022~\cite{van2022biomechanics} have unreported sex and are fitted with a SKEL sex model chosen by height (female below $1.683$\,m, male otherwise). Statistics and benchmark results for the rare classes rest on few trials and should be read with this caveat. Per-source counts are listed in Table~\ref{tab:dataset_composition}.}

ForceBody aggregates motion trials from ten public motion-capture datasets included in AddBiomechanics~\cite{werling2024addbiomechanics}. After applying the quality-based filter (Appendix~\ref{app:filtering}), the released dataset spans 140 subjects, \red{10,386} trials, and approximately 26.9 hours of recorded motion. Table~\ref{tab:dataset_composition} summarizes the per-source composition: trial and subject counts, the male/female/unreported breakdown, mean$\pm$std subject height and mass, and the dominant activity type.

\begin{table}[h]
\centering
\small
\caption{Per-dataset composition of \textbf{ForceBody}. Subjects refers to unique individuals; each subject may contribute multiple recording sessions. M/F denotes male/female; U = unreported.}
\label{tab:dataset_composition}
\begin{tabular}{lrrcccc}
\toprule
Source & Trials & Subj. & M/F/U & Height (m) & Mass (kg) & Activity \\
\midrule
Carter2023~\cite{carter2024full}
  & $7{,}006$ & $50$ & 25/25/0 & \red{$1.73{\pm}0.08$} & \red{$65.8{\pm}10.6$} & \red{Running / walk} \\
vanderZee2022~\cite{van2022biomechanics}
  & $1{,}089$ & $10$ & 0/0/10  & \red{$1.76{\pm}0.11$} & \red{$73.5{\pm}15.0$} & Walking \\
Moore2015~\cite{moore2015elaborate}
  & $1{,}019$ & $12$ & 8/4/0   & \red{$1.73{\pm}0.09$} & \red{$71.4{\pm}11.9$} & Walking \\
\red{Tiziana2019~\cite{lencioni2019human}}
  & \red{$619$} & \red{$25$} & \red{14/11/0} & \red{$1.71{\pm}0.11$} & \red{$69.6{\pm}14.2$} & \red{Walking} \\
Han2023~\cite{han2023groundlink}
  & \red{$240$} & $7$  & 3/4/0   & \red{$1.72{\pm}0.14$} & \red{$66.4{\pm}13.6$} & \red{Mixed} \\
Uhlrich2023~\cite{uhlrich2023opencap}
  & \red{$160$} & \red{$10$} & \red{4/6/0} & \red{$1.74{\pm}0.12$} & \red{$69.2{\pm}11.6$} & \red{Walk / jump} \\
Fregly2012~\cite{fregly2012grand}
  & $78$      & $4$  & 3/1/0   & \red{$1.71{\pm}0.06$} & \red{$71.3{\pm}6.5$}  & Walking \\
Falisse2017~\cite{falisse2016emg}
  & \red{$70$} & \red{$11$} & \red{7/4/0} & \red{$1.76{\pm}0.06$} & \red{$70.0{\pm}6.4$}  & Sit-to-stand / walk \\
Li2021~\cite{li2021well}
  & $65$      & $1$  & 1/0/0   & $1.83$          & $85.7$          & \red{Walking} \\
Hamner2013~\cite{hamner2013muscle}
  & $40$      & $10$ & 10/0/0  & $1.74{\pm}0.05$ & \red{$71.6{\pm}6.3$}  & Running \\
\midrule
Total
  & \red{$10{,}386$} & $140$ & \red{75/55/10} & $1.73{\pm}0.09$ & $68.8{\pm}11.5$ & \\
\bottomrule
\end{tabular}
\end{table}


\section{Fitting Pipeline}
\label{app:fitting}

\begin{algorithm}[h]
\caption{SKEL fitting.}
\label{alg:fitting}
\begin{algorithmic}[1]
\Require AddB joint positions $\mathbf{J}^{\text{GT}}_{1:T}$, marker positions $\mathbf{p}^{\text{obs}}_{1:T}$, reference angles $\bm{\theta}^{\text{ref}}_{1:T}$ with mask $\mathbf{m}$
\Ensure SKEL shape $\bm{\beta}$ and per-frame pose/translation $\{\bm{\theta}_t, \mathbf{t}_t\}_{t=1}^{T}$
\\\hrulefill
\State \textbf{Stage 1 — Shape estimation}
\State $\bar{\bm{\ell}}^{\text{GT}} \gets$ \red{median over the subject's trials} of \red{per-trial mean} AddB bone \red{lengths}
\State $\bm{\beta} \gets \displaystyle\arg\min_{\bm{\beta}} \sum_i w_i\,\bigl(\ell_i^{\text{SKEL}}(\bm{\beta}) - \bar{\ell}_i^{\text{GT}}\bigr)^2$ \Comment{Adam, \red{$1{,}000$} iters, best of 6 starts}
\State Freeze $\bm{\beta}$
\\\hrulefill
\State \textbf{Stage 2 — Pose optimization}
\For{$k = 1, \dots, K_\text{iter}$}
 \State $\mathcal{L}_{\text{pos}} \gets \lambda_\text{j}\mathcal{L}_\text{joint} + \lambda_\text{b}\mathcal{L}_\text{bone} + \lambda_\text{w}\mathcal{L}_\text{width} + \lambda_\text{f}\mathcal{L}_\text{foot} + \lambda_\text{m}\mathcal{L}_\text{marker}$
 \State $\mathcal{L}_{\text{reg}} \gets \lambda_\text{p}\|\bm{\theta}\|^2 + \lambda_\text{s}\|\bm{\theta}_\text{spine}\|^2 + \lambda_\text{sc}\|\bm{\theta}_\text{scap}\|^2 + \lambda_\text{h}\|\bm{\theta}_\text{hum}\|^2$
 \State $\mathcal{L}_{\text{temp}} \gets \dfrac{\lambda_\text{v}}{T}\sum_t\|\dot{\bm{\theta}}_t\|^2 + \dfrac{\lambda_\text{a}}{T}\sum_t\|\ddot{\bm{\theta}}_t\|^2$
  \State $\mathcal{L}_{\text{q-match}} \gets \red{\dfrac{1}{T}\sum_t} \bigl\|(\bm{\theta}_t - \bm{\theta}^{\text{ref}}_t)\odot\mathbf{m}\bigr\|^2$
 \State $\mathcal{L} \gets \mathcal{L}_{\text{pos}} + \mathcal{L}_{\text{reg}} + \mathcal{L}_{\text{temp}} + \lambda_\text{q}\mathcal{L}_{\text{q-match}}$
 \State $\{\bm{\theta}_t, \mathbf{t}_t\} \gets \text{Adam step on } \nabla\mathcal{L}$
 \State \textbf{break} if early-stop patience exceeded
\EndFor
\State \red{Foot-only} fine-tuning ($400$ iters)
\State \Return $\bm{\beta},\, \{\bm{\theta}_t, \mathbf{t}_t\}_{t=1}^{T}$
\end{algorithmic}
\end{algorithm}

\paragraph{Preprocessing.} Source datasets span native sampling rates of $61$--\red{$250$}\,Hz. \red{We filter the kinematics with the same 2nd-order 30\,Hz zero-lag Butterworth filter that AddBiomechanics~\cite{werling2024addbiomechanics} applies in its next pass. About $98\%$ of the trials are sampled at about $100$\,Hz (99 to 101\,Hz), and the others keep their native rate.} We use kinematics from the inverse-kinematics pass only, excluding the force-based kinematics refinement of AddBiomechanics, to prevent force labels from leaking into kinematic inputs~\cite{tan2026gaitdynamics}.

\paragraph{Two-stage fitting.} For each trial we recover SKEL parameters $(\bm{\beta}, \{\bm{\theta}_t, \mathbf{t}_t\}_{t=1}^T)$ in two sequential stages (Algorithm~\ref{alg:fitting}). \textbf{Stage 1} estimates one shape $\bm{\beta}\in\mathbb{R}^{10}$ per subject by matching \red{eight bilateral} bone lengths \red{(pelvis} to \red{hip,} \red{thigh,} \red{shank,} forearm). The target \red{is} \red{the} \red{median} \red{over} \red{the} \red{subject's trials} of \red{each} \red{trial's} \red{mean} \red{bone length.} The shin bone pair is weighted $10\times$ because SKEL's shape space couples thigh and shin dimensions; without the upweighting, the optimizer trades shin error for thigh error, producing knee-level fitting artifacts. \textbf{Stage 2} fixes $\bm{\beta}$ and optimizes the per-frame pose $\bm{\theta}_t\in\mathbb{R}^{46}$ and translation $\mathbf{t}_t\in\mathbb{R}^{3}$ under the composite loss

\begin{equation}
\mathcal{L} \;=\; \mathcal{L}_{\text{pos}} \;+\; \mathcal{L}_{\text{reg}} \;+\; \mathcal{L}_{\text{temp}} \;+\; \lambda_\text{q}\,\mathcal{L}_{\text{q-match}}.
\label{eq:supp_pose_loss}
\end{equation}

\paragraph{Positional loss.} The positional term $\mathcal{L}_{\text{pos}}$ aggregates five sub-terms:
\begin{equation}
\mathcal{L}_{\text{pos}} \;=\; \lambda_\text{j}\,\mathcal{L}_\text{joint} \;+\; \lambda_\text{b}\,\mathcal{L}_\text{bone} \;+\; \lambda_\text{w}\,\mathcal{L}_\text{width} \;+\; \lambda_\text{f}\,\mathcal{L}_\text{foot} \;+\; \lambda_\text{m}\,\mathcal{L}_\text{marker},
\end{equation}
where $\mathcal{L}_\text{joint}$ is a \red{per-joint-weighted} Huber loss ($\delta{=}50$\,mm) over the 15 directly matched joints, $\mathcal{L}_\text{bone}$ penalizes squared bone-length deviation across 12 pairs, $\mathcal{L}_\text{width}$ aligns shoulder width, $\mathcal{L}_\text{foot}$ enforces vertical alignment of foot joints (critical for downstream GRF consistency), and $\mathcal{L}_\text{marker}$ matches BSM mesh vertices to the observed optical markers.

\paragraph{Regularization and temporal smoothness.} The regularizer $\mathcal{L}_{\text{reg}}$ penalizes unnatural rotations in under-constrained regions where AddB provides no direct anatomical anchor:
\begin{equation}
\mathcal{L}_{\text{reg}} \;=\; \lambda_\text{p}\,\|\bm{\theta}\|^2 + \lambda_\text{s}\,\|\bm{\theta}_\text{spine}\|^2 + \lambda_\text{sc}\,\|\bm{\theta}_\text{scap}\|^2 + \lambda_\text{h}\,\|\bm{\theta}_\text{hum}\|^2.
\end{equation}
The temporal term $\mathcal{L}_{\text{temp}}$ enforces smoothness via finite differences,
\begin{equation}
\mathcal{L}_{\text{temp}} \;=\; \frac{\lambda_\text{v}}{T}\sum_t \|\dot{\bm{\theta}}_t\|^2 + \frac{\lambda_\text{a}}{T}\sum_t \|\ddot{\bm{\theta}}_t\|^2.
\end{equation}

\paragraph{DOF-matching prior.} The optional pose-matching term $\mathcal{L}_\text{q-match}$ softly anchors SKEL DOFs to a per-frame reference pose $\bm{\theta}^{\text{ref}}_t$, which is obtained either by (i) directly estimating lower-body SKEL DOFs from rotation matrices computed against the AddB skeleton, or (ii) by mapping AddB Rajagopal joint angles to SKEL DOF indices via fixed scale factors. The term is gated by a per-DOF mask $\mathbf{m}$ that selects only the DOFs for which a reference is available, and is \red{active} \red{with} \red{a constant weight} from \red{the} \red{first} \red{iteration.} \red{Each} \red{DOF is further weighted by the confidence of its reference.}
\begin{equation}
\mathcal{L}_\text{q-match} \;=\; \frac{1}{T}\sum_t \bigl\|(\bm{\theta}_t - \bm{\theta}^{\text{ref}}_t)\odot\mathbf{m}\bigr\|^2,
\end{equation}
We set $\lambda_\text{q}{=}0.1$, so this term contributes a small fraction of the total loss and primarily serves to stabilize lower-body and mapped DOFs where the AddB reference is available; under-constrained DOFs without an AddB counterpart (spine, scapula, humerus) are instead regularized by $\mathcal{L}_\text{reg}$.

\paragraph{Hyperparameters.} \red{Stage} \red{1} \red{uses} Adam with \red{a constant} learning rate \red{of} \red{$0.01$} for \red{$1{,}000$} iterations \red{from} \red{six} \red{starting} \red{points} \red{per} \red{subject and keeps the start with the lowest loss.} Stage 2 runs $K_\text{iter}{=}500$ iterations \red{in} \red{two} \red{halves with cosine-annealed Adam (pose learning rates $0.02$} and \red{then} \red{$0.01$)} \red{and} early stopping at patience $80$. The Huber loss uses $\delta{=}50$\,mm. \red{A} pelvis-orientation \red{loss} \red{aligns} the \red{SKEL} pelvis \red{axes} \red{with} the \red{AddB} \red{pelvis} \red{orientation. The marker term is part of} the main \red{objective,} \red{and} a foot-only fine-tuning ($400$ iterations) \red{cleans} up foot-ground \red{contact} \red{afterwards.} Loss weights are listed in Table~\ref{tab:supp_loss_weights}.

\begin{table}[h]
\centering
\small
\caption{Stage 2 loss weights.}
\label{tab:supp_loss_weights}
\begin{tabular}{lcc}
\toprule
Term & Symbol & Weight \\
\midrule
Joint position & $\lambda_\text{j}$ & $1.0$ \\
Bone length & $\lambda_\text{b}$ & $1.0$ \\
Shoulder width & $\lambda_\text{w}$ & $2.0$ \\
Foot height & $\lambda_\text{f}$ & $15.0$ \\
Marker & $\lambda_\text{m}$ & $10.0$ \\
Pose reg. & $\lambda_\text{p}$ & $0.01$ \\
Spine reg. & $\lambda_\text{s}$ & $0.3$ \\
Scapula reg. & $\lambda_\text{sc}$ & $0.05$ \\
Humerus reg. & $\lambda_\text{h}$ & $0.05$ \\
Velocity & $\lambda_\text{v}$ & $0.01$ \\
Acceleration & $\lambda_\text{a}$ & $0.1$ \\
DOF match & $\lambda_\text{q}$ & $0.1$ \\
\bottomrule
\end{tabular}
\end{table}


\section{Pseudo-code for per-frame torque uncertainty}
\label{app:mc_algo}

For every trial of \textbf{ForceBody}, we compute the per-frame, per-joint torque uncertainty $\sigma_{\tau}$ by Monte Carlo sampling through the same OpenSim inverse-dynamics (ID) solver that produced the ground-truth torques. We draw $N{=}1000$ band-limited Gaussian perturbations of the AddB joint angles using an FFT-based generator with RMS $\sigma_q{=}1.6^{\circ}$ and a $6$\,Hz lowpass cutoff. Noise is injected only on the 20 DOFs spanning the 14 major joints with a direct SKEL correspondence (bilateral hip, knee, ankle, MTP, elbow, radioulnar, wrist). We add the position noise to the released GT pose and propagate it to velocity and acceleration by \red{backward} finite differencing, leaving the GT $\dot{\bm{q}}$ and $\ddot{\bm{q}}$ as the baseline so that the resulting torque variance is attributable to $\sigma_q$ alone. We then run AddB inverse dynamics on the perturbed kinematics with the measured GRF held fixed, and aggregate the per-DOF mean $\mu_{\tau}$ and standard deviation $\sigma_{\tau}$ across the $N$ samples. The procedure is summarized in Algorithm~\ref{alg:mc_validation}.

\begin{algorithm}[t]
\caption{Per-frame torque uncertainty via Monte Carlo sampling.}
\label{alg:mc_validation}
\begin{algorithmic}[1]
\Require AddB kinematics $\bm{q}_{1:T},\, \dot{\bm{q}}_{1:T},\, \ddot{\bm{q}}_{1:T}$;
         measured GRF $\bm{\lambda}_{1:T}$;
         direct-DOF mask $\mathcal{M}$;
         pose RMS $\sigma_q{=}1.6^{\circ}$;
         cutoff $f_c{=}6$\,Hz;
         samples $N{=}1000$
\For{$n = 1, \dots, N$}
  \State $\bm{\epsilon}^{(n)} \gets \mathrm{BandLimitedGaussian}(T,\, \sigma_q,\, f_c)$
         \Comment{zero outside $\mathcal{M}$}
  \State $\dot{\bm{\epsilon}}^{(n)},\, \ddot{\bm{\epsilon}}^{(n)} \gets \mathrm{BackwardDiff}(\bm{\epsilon}^{(n)})$
  \State $\tilde{\bm{\tau}}^{(n)} \gets
          \mathrm{ID}\bigl(\bm{q}{+}\bm{\epsilon}^{(n)},\, \dot{\bm{q}}{+}\dot{\bm{\epsilon}}^{(n)},\, \ddot{\bm{q}}{+}\ddot{\bm{\epsilon}}^{(n)},\, \bm{\lambda}\bigr)$
\EndFor
\State $\mu_{\tau,t},\, \sigma_{\tau,t} \gets
        \mathrm{mean/std}\bigl\{\tilde{\bm{\tau}}_t^{(n)}\bigr\}_{n=1}^{N}$
        \Comment{per frame $t$, per DOF}
\State \Return $\{\mu_{\tau,t},\, \sigma_{\tau,t}\}_{t=1}^{T}$
\end{algorithmic}
\end{algorithm}

\begin{figure}[!t]
\centering
\includegraphics[width=\textwidth]{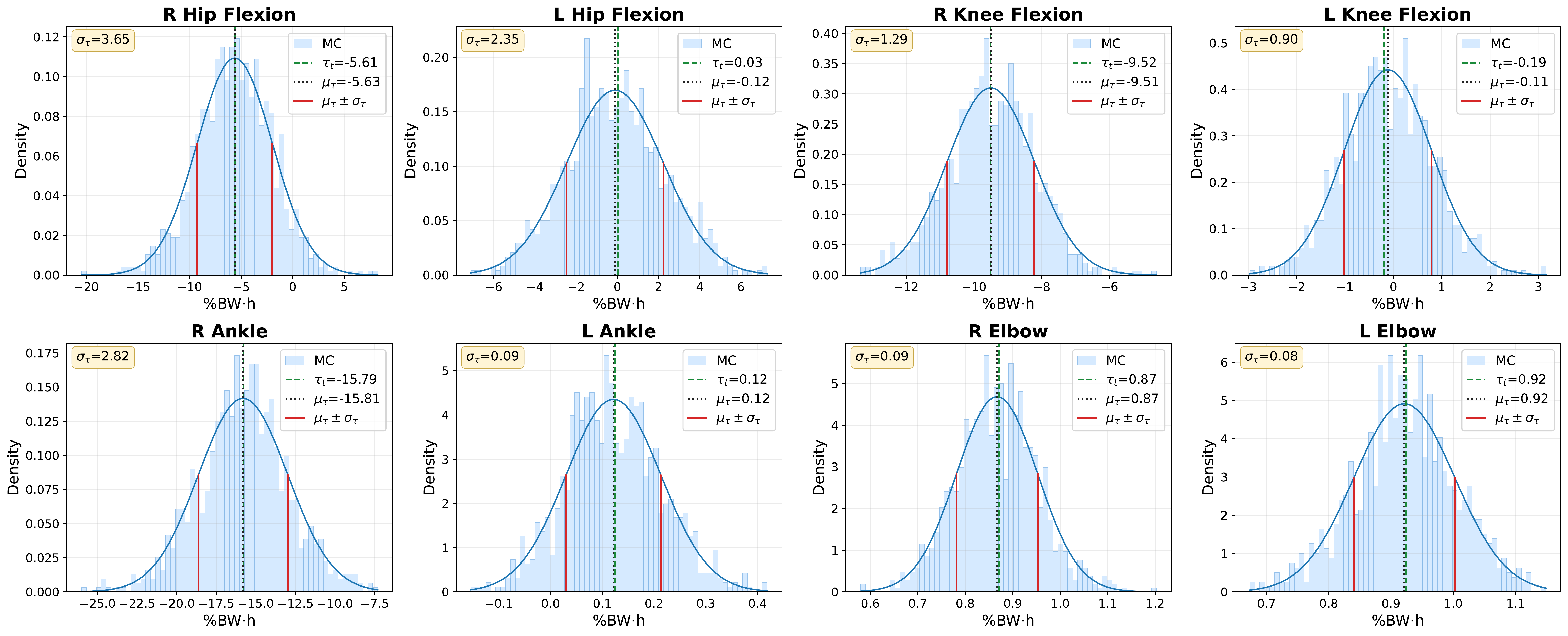}
\caption{\textbf{Per-joint torque uncertainty for one representative frame
(extended).} Same trial and frame as main paper, extended to both right and left sides.}
\label{fig:per_joint_sigma_supp}
\end{figure}

\section{Per-Sample Torque Uncertainty: Statistics}
\label{app:uncertainty_stats}

This section breaks down the released $\bm{\sigma}_{\tau}$ across joints, activities, and source datasets on the \red{$8{,}732$} released trials. All values are mass-normalized (Nm/kg) and reported as the cross-trial median of the per-trial median $\sigma_{\tau}$.

\paragraph{Per joint.} Table~\ref{tab:supp_sigma_per_joint} reports $\bm{\sigma}_{\tau}$ on the 14 direct joints. Hip and knee carry the widest envelopes ($0.47$ and $0.21$\,Nm/kg) due to their long lever arms. The toe joint (MTP) and the distal forearm joints (radioulnar, wrist) carry the tightest ($\sim\!10^{-3}$\,Nm/kg).

\begin{table}[h]
\centering
\small
\red{\setlength{\tabcolsep}{4pt}}
\begin{minipage}[t]{0.48\linewidth}
\red{\centering}
\caption{Per-joint $\bm{\sigma}_{\tau}$ across the 14 direct joints. Left and right averaged. \textit{Trial-Med} is the cross-trial median of each per-trial median $\sigma_{\tau}$. P25 and P75 are the inter-quartile range.}
\label{tab:supp_sigma_per_joint}
\resizebox{\linewidth}{!}{\begin{tabular}{lccc}
\toprule
Joint & Trial-Med [Nm/kg] & P25 & P75 \\
\midrule
Hip & $0.466$ & $0.448$ & $0.494$ \\
Knee & $0.211$ & $0.200$ & $0.224$ \\
Ankle & $0.116$ & $0.100$ & $0.125$ \\
MTP & $0.0009$ & $0.0008$ & $0.0010$ \\
Elbow & $0.015$ & $0.013$ & $0.016$ \\
Radioulnar & $0.0004$ & $0.0004$ & $0.0004$ \\
Wrist & $0.0017$ & $0.0016$ & $0.0018$ \\
\bottomrule
\end{tabular}}
\end{minipage}\hfill
\begin{minipage}[t]{0.48\linewidth}
\red{\centering}
\caption{$\bm{\sigma}_{\tau}$ broken down by activity class. \textit{Sigma-Med} is the cross-trial median of the per-trial median $\sigma_{\tau}$ pooled across the 14 direct joints. \textit{Rel-Med} is the cross-trial median of $\sigma_{\tau}/|\tau|$.}
\label{tab:supp_sigma_per_activity}
\resizebox{\linewidth}{!}{\begin{tabular}{lcrc}
\toprule
\red{Activity} & \red{Trials} & \red{Sigma-Med [Nm/kg]} & \red{Rel-Med} \\
\midrule
\red{Jumping} & \red{$52$} & \red{$0.017$} & \red{$0.36$} \\
\red{Running} & \red{$4{,}959$} & \red{$0.013$} & \red{$0.50$} \\
\red{Sit-to-stand} & \red{$29$} & \red{$0.018$} & \red{$0.78$} \\
\red{Others} & \red{$148$} & \red{$0.018$} & \red{$0.88$} \\
\red{Walking} & \red{$3{,}310$} & \red{$0.016$} & \red{$0.92$} \\
\red{Stairs} & \red{$4$} & \red{$0.015$} & \red{$1.26$} \\
\red{Standing} & \red{$230$} & \red{$0.017$} & \red{$1.28$} \\
\bottomrule
\end{tabular}}
\end{minipage}
\end{table}

\paragraph{Per activity.} Table~\ref{tab:supp_sigma_per_activity} reports $\bm{\sigma}_{\tau}$ and the relative envelope $\sigma_{\tau}/|\tau|$ across activity classes. Dynamic activities (\red{jumping, running}) carry the smallest relative envelopes (rel-med \red{$0.36$ for jumping, $0.50$ for running}). Near-static activities (\red{standing}) carry the largest (\red{$1.28$}). Trials whose activity label is missing or does not fall into one of the named classes are pooled in \textit{Others}. The absolute envelope $\sigma_{\tau}$ is roughly comparable across activities ($0.013$--\red{$0.018$}\,Nm/kg), so the relative trend is driven mostly by the size of the underlying torque signal $|\tau|$ rather than by changes in $\sigma_{\tau}$ itself.

\paragraph{Per source.} Table~\ref{tab:supp_sigma_per_source} reports the same statistics broken down by source dataset. Running-dominated sources (\red{Hamner2013}, Carter2023) sit at the low end of the relative envelope (\red{$0.45$ and $0.53$}). Sources dominated by slow or quasi-static motion (\red{Fregly2012, Falisse2017 sit-to-stand}) sit at the high end (\red{$0.98$ and $1.06$}).

\begin{table}[h]
\centering
\small
\caption{$\bm{\sigma}_{\tau}$ broken down by source dataset. \textit{Sigma-Med} and \textit{Rel-Med} follow Table~\ref{tab:supp_sigma_per_activity}.}
\label{tab:supp_sigma_per_source}
\begin{tabular}{lcrc}
\toprule
Source        & Trials & Sigma-Med [Nm/kg] & Rel-Med \\
\midrule
\red{Hamner2013}    & \red{$40$}      & \red{$0.018$} & \red{$0.45$} \\
\red{Carter2023}    & \red{$6{,}385$} & \red{$0.014$} & \red{$0.53$} \\
\red{Tiziana2019}   & \red{$46$}      & \red{$0.014$} & \red{$0.59$} \\
\red{Li2021}        & \red{$19$}      & \red{$0.019$} & \red{$0.79$} \\
\red{Uhlrich2023}   & \red{$112$}     & \red{$0.017$} & \red{$0.79$} \\
\red{vanderZee2022} & \red{$1{,}020$} & \red{$0.016$} & \red{$0.83$} \\
\red{Moore2015}     & \red{$849$}     & \red{$0.016$} & \red{$0.85$} \\
\red{Han2023}       & \red{$176$}     & \red{$0.018$} & \red{$0.91$} \\
\red{Fregly2012}    & \red{$62$}      & \red{$0.017$} & \red{$0.98$} \\
\red{Falisse2017}   & \red{$23$}      & \red{$0.016$} & \red{$1.06$} \\
\bottomrule
\end{tabular}
\end{table}

\paragraph{Joint-by-activity relative uncertainty.} Table~\ref{tab:supp_sigma_joint_activity} cross-tabulates $\sigma_{\tau}/|\tau|$ by joint and activity on the 14 direct joints. Hip is the joint whose relative uncertainty changes the most across activities, dropping to \red{$0.93$} in running and rising to \red{$4.55$ in standing}, where hip torque is dominated by quasi-static body-weight support and the kinematic-noise contribution becomes a larger fraction of the signal. Knee shows a similar pattern but on a smaller scale, while distal joints (MTP, radioulnar, wrist) stay closer across activities.

\begin{table}[h]
\centering
\small
\caption{Joint-by-activity median relative uncertainty $\sigma_{\tau}/|\tau|$ on the 14 direct joints.}
\label{tab:supp_sigma_joint_activity}
\setlength{\tabcolsep}{4pt}
\begin{tabular}{lccccccc}
\toprule
Joint      & Run  & Jump & Sit-to-stand & Others & Walk & Stair & Standing \\
\midrule
Hip        & \red{$0.93$} & \red{$1.30$} & \red{$1.90$} & \red{$2.12$} & \red{$1.81$} & \red{$2.21$} & \red{$4.55$} \\
Knee       & \red{$0.54$} & \red{$0.23$} & \red{$0.44$} & \red{$0.64$} & \red{$0.99$} & \red{$0.80$} & \red{$2.19$} \\
Ankle      & \red{$0.28$} & \red{$0.27$} & \red{$0.43$} & \red{$0.35$} & \red{$0.27$} & \red{$0.27$} & \red{$0.40$} \\
MTP        & \red{$0.53$} & \red{$0.79$} & \red{$0.94$} & \red{$0.88$} & \red{$0.67$} & \red{$0.80$} & \red{$0.84$} \\
Elbow      & \red{$0.41$} & \red{$0.41$} & \red{$0.58$} & \red{$1.20$} & \red{$0.95$} & \red{$1.36$} & \red{$0.97$} \\
Radioulnar & \red{$0.57$} & \red{$0.35$} & \red{$0.68$} & \red{$0.67$} & \red{$0.77$} & \red{$1.26$} & \red{$1.85$} \\
Wrist      & \red{$0.48$} & \red{$0.34$} & \red{$0.78$} & \red{$0.98$} & \red{$1.08$} & \red{$2.14$} & \red{$1.35$} \\
\bottomrule
\end{tabular}
\end{table}

\section{Quality-based filtering of \textbf{ForceBody}}
\label{app:filtering}

The raw trial data can contain heavy-tailed outlier trials, typically caused by very fast running, sharp cuts, or abnormal body shapes. To ensure data quality, all quantitative results in this paper are computed on a quality-filtered subset of \textbf{ForceBody}. Specifically, we consider the threshold suggested in~\cite{hicks2015my} and define a single hard cutoff. This leads to a subset with root-relative MPJPE is at or below the $80$th percentile of the unfiltered baseline. This filtering improves data quality while preserving all $10$ source datasets and all activity categories. The resulting subset contains \red{$10{,}386$} trials ($80.0\%$ of $12{,}986$), \red{$9{,}689{,}808$} frames ($\sim 26.9$ hours), and $140$ of the original $143$ unique subjects ($97.9\%$).


\section{Fitting Quality of ForceBody}\label{app:fitting_quality}

This section provides the full breakdown of fitting quality omitted from
the main paper, evaluated on all \red{$10{,}386$} released trials. Per-joint
MPJPE values are reported on the $15$ joints with a direct
correspondence between the AddB Rajagopal skeleton and the SKEL
kinematic tree, with left and right sides averaged.

\textbf{Aggregate metrics.} We report two complementary metric groups
in Table~\ref{tab:supp_fitting_quality}, top. \textbf{Positional errors}
compare predicted SKEL joint centers to the AddB joint centers. MPJPE
is the mean per-joint Euclidean distance in world coordinates, and
PA-MPJPE applies a per-frame Procrustes alignment \red{that removes global rotation, translation, and scale before measuring the error.} \textbf{Physical-plausibility errors}
measure how realistic the resulting motion is. Per-frame velocity and
acceleration errors (mm/f, mm/f$^2$) capture trajectory smoothness, and
foot sliding and ground penetration (both in mm) capture foot-contact
artifacts. Table~\ref{tab:fitting_quality_full} shows the comprehensive analysis on the fitting quality.

\textbf{Per-joint breakdown.} Table~\ref{tab:supp_fitting_quality},
bottom, reports per-joint MPJPE on the $15$ direct joints. Lower-body
joints (pelvis, hip, knee, ankle, MTP) average \red{$16.6$}\,mm. Upper-body
joints (elbow, forearm, wrist) average \red{$29.0$}\,mm. The upper-body
errors are larger across all three joints, consistent with the wider
operating range of the arms in everyday motion and the sparser optical
marker coverage of the upper body in the source datasets.

\begin{table}[t]
\centering
\small
\caption{\textbf{Top}: Aggregate fitting quality of \textbf{ForceBody}.
\textbf{Bottom}: per-joint MPJPE on the $15$ directly matched joints,
left and right averaged. Pelv.=Pelvis, Kne.=Knee, Ank.=Ankle,
MTP=Metatarsophalangeal joint, F.a.=Forearm (radioulnar), Wri.=Wrist,
Elb.=Elbow.}
\label{tab:supp_fitting_quality}
\begin{tabular*}{0.85\linewidth}{@{\extracolsep{\fill}} l cc cccc @{}}
\toprule
 & \multicolumn{2}{c}{Positional Err.} & \multicolumn{4}{c}{Physical-plausibility Err.} \\
\cmidrule(lr){2-3} \cmidrule(lr){4-7}
 & MPJPE & PA-MPJPE & Vel.\ err.\ & Acc.\ err.\ & Foot slid.\ & Ground pen.\ \\
 & [mm]  & [mm]     & [mm/f]      & [mm/f$^2$]   & [mm]        & [mm]         \\
\midrule
\textbf{ForceBody} & \red{$21.6$} & \red{$21.2$} & \red{$0.89$} & \red{$0.24$} & \red{$1.23$} & \red{$1.17$} \\
\midrule
\midrule
\end{tabular*}%
\\[-\arrayrulewidth]
\begin{tabular*}{0.85\linewidth}{@{\extracolsep{\fill}} l ccccc ccc @{}}
 & \multicolumn{5}{c}{Lower body (Avg.\ \red{16.6})} & \multicolumn{3}{c}{Upper body (Avg.\ \red{29.0})} \\
\cmidrule(lr){2-6} \cmidrule(lr){7-9}
 & MTP & Ank. & Kne. & Hip & Pelv. & Wri. & F.a. & Elb. \\
\midrule
Mean & \red{$15.9$} & \red{$17.0$} & \red{$20.5$} & \red{$13.8$} & \red{$15.9$} & \red{$29.3$} & \red{$28.4$} & \red{$29.3$} \\
\bottomrule
\end{tabular*}
\end{table}

\begin{figure}[!t]
  \centering
  \includegraphics[width=0.95\linewidth]{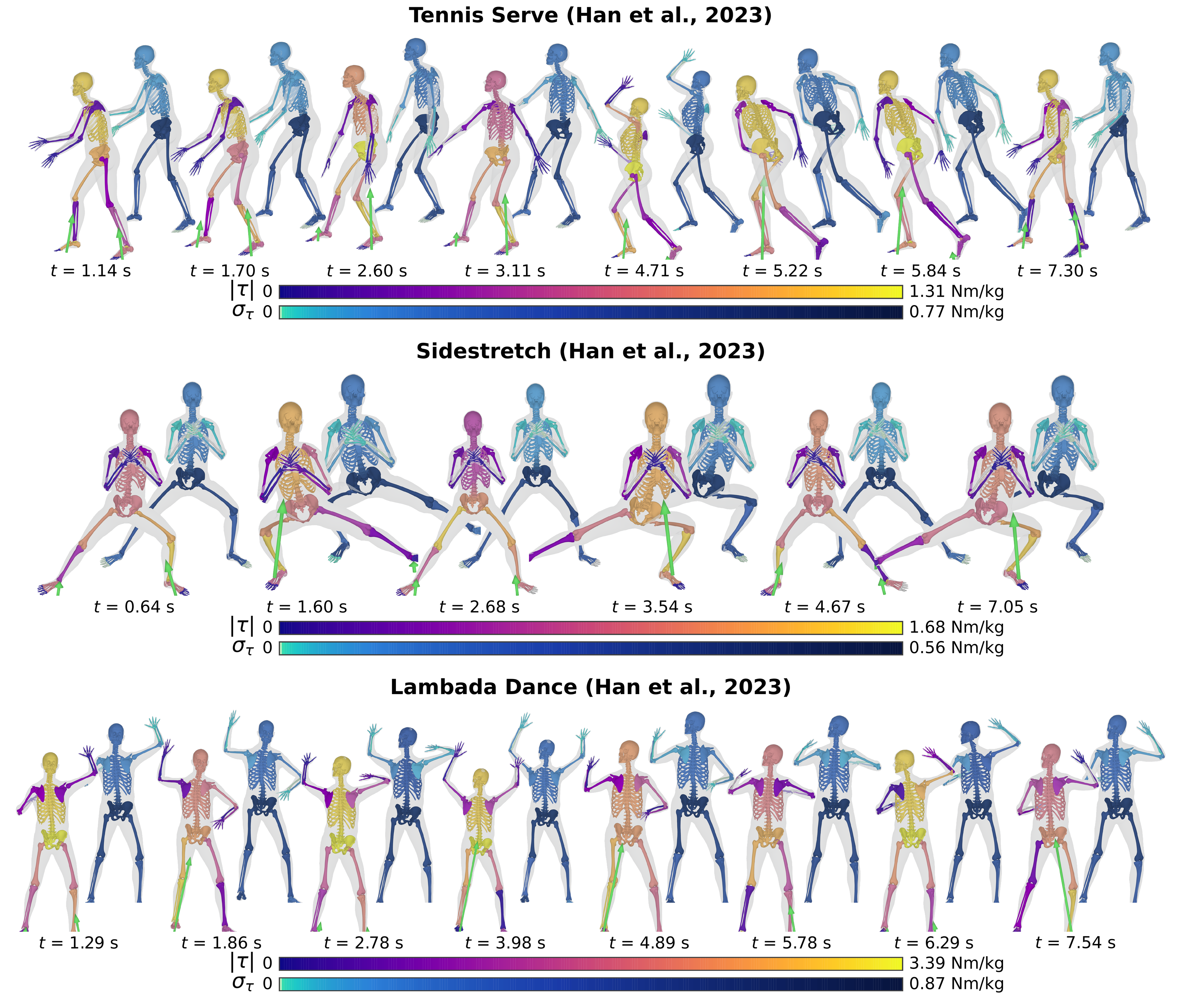}
  \caption{SKEL fit across time on three trials from \cite{han2023groundlink}, top to bottom Tennis Serve, Sidestretch, and Lambada Dance. Each body is shown twice, with bones colored by
$|\tau|$ (front) and by $\sigma_\tau$ (back), both in Nm/kg (colorbars below each
row). Green arrows are measured GRF.}

  \label{fig:temporal_strips}
\end{figure}

\paragraph{Additional qualitative examples.}
Figure~\ref{fig:temporal_strips} shows the SKEL fit at evenly spaced frames on three dynamic trials from \cite{han2023groundlink}, with per-frame torque magnitude $|\bm{\tau}_t|$ and Monte Carlo uncertainty $\sigma_{\bm{\tau},t}$ overlaid on the 14 directly matched joints. The fit tracks the underlying motion across high-dynamic, sustained, and rhythmic regimes. 

\section{Joint Torque Training Details}\label{app:training}

This section provides the training details for the joint torque baselines reported in the main paper.

\paragraph{Targets and masking.}
The supervision target at each frame is the AddB inverse-dynamics torque vector $\bm{\tau}_t \in \mathbb{R}^{37}$ on the Rajagopal skeleton. \red{The} loss \red{covers all 37 DOFs. We report errors on} the 14 anatomical joints (20 DOFs) with a direct SKEL correspondence (bilateral hip, knee, ankle, MTP, elbow, radioulnar, and wrist), matching the joint set used for $\sigma_\tau$ in Section~\ref{app:uncertainty_stats}. We additionally mask any frame whose GRF coverage is invalid in the AddB labels.

\begin{table}[!t]
\centering
\small
\setlength{\tabcolsep}{6pt}
\caption{Full distribution of per-trial fitting-quality metrics across
the quality-filtered subset of \textbf{ForceBody} (\red{$10{,}386$} trials).}
\label{tab:fitting_quality_full}
\begin{tabular}{lrrrr}
\toprule
Metric & Mean & Median & P$5$ & P$95$ \\
\midrule
MPJPE (mm)                    & \red{$21.6$} & \red{$22.7$} & \red{$10.5$} & \red{$32.4$} \\
PA-MPJPE (mm)                 & \red{$21.2$} & \red{$22.4$} & \red{$10.1$} & \red{$32.0$} \\
Velocity error (mm/f)         & \red{$0.89$} & \red{$0.97$} & \red{$0.19$} & \red{$1.63$} \\
Acceleration error (mm/f$^2$) & \red{$0.24$} & \red{$0.24$} & \red{$0.07$} & \red{$0.44$} \\
Foot sliding (mm)             & \red{$1.23$} & $0.00$ & $0.00$ & \red{$9.26$} \\
Ground penetration (mm)       & \red{$1.17$} & \red{$0.20$} & $0.00$ & \red{$7.34$} \\
\bottomrule
\end{tabular}
\end{table}

\paragraph{Per-DOF $z$-score normalization.}
Joint torque magnitudes vary by two orders of magnitude across the body, from $\sim\!0.01$\,Nm/kg at the wrist and toe joints to $\sim\!1$\,Nm/kg at the hip. Training a single model on raw torque values is therefore dominated by the high-magnitude lower-body joints. To balance the loss across joints, we normalize each DOF independently by its training-set mean $\mu_d$ and standard deviation $\sigma_d$,
\begin{equation}
\tilde{\tau}_{t,d} \;=\; \frac{\tau_{t,d} - \mu_d}{\sigma_d}, \quad d \in \{1, \dots, 37\},
\end{equation}
so that every DOF contributes comparable magnitude to the loss. The model predicts $\tilde{\tau}$ in normalized space, and we de-normalize at evaluation time to report errors in physical Nm/kg.

\paragraph{Loss.}
The training loss is the mean-squared error over \red{all} \red{37 normalized} DOFs and \red{the} valid frames in a batch,
\begin{equation}
\mathcal{L}_\tau \;=\; \frac{1}{37\sum_{t} v_t}\sum_{t,d} v_t \cdot \bigl(\hat{\tilde{\tau}}_{t,d} - \tilde{\tau}_{t,d}\bigr)^2,
\end{equation}
where $v_t \in \{0, 1\}$ is the per-frame validity flag.

\paragraph{Optimization.}
All baselines are optimized with Adam at learning rate $2{\times}10^{-4}$, batch size $128$, and sequence length $256$ frames ($2.56$\,s at $100$\,Hz). We \red{use $1{,}000$ warm-up steps, decay the learning rate by a factor of $0.05$ after $200$k steps, and stop after $80$ epochs without improvement on the held-out split. The held-out split is the validation subjects (12 training subjects, 1{,}098 trials). We} train each model on a single NVIDIA L40S \red{GPU.}

\paragraph{Architectures.}
We benchmark six sequence models that span the standard families used in human dynamics regression. The MLP follows the AddBiomechanics baseline of \cite{werling2024addbiomechanics}, a per-frame 2-layer feedforward network with hidden width $512$ and no temporal context. Conv1D is a fully convolutional residual network with kernel size $3$, $256$ channels, and $8$ residual blocks. LSTM is a single-layer bidirectional recurrent network with hidden size $256$, and GRU is a $4$-layer unidirectional recurrent network with hidden size $512$. Mamba2 follows the MInT configuration of \cite{schneider2024muscles} with $d_{\text{model}}{=}512$ and $8$ stacked Mamba2 blocks. The Transformer is a $16$-layer encoder with $16$ attention heads, hidden size $256$, and feedforward width $512$. Conv1D, LSTM, and Transformer also follow the MInT configurations. All models \red{take} the same $153$-dim SKEL state vector \red{as input} and \red{predict} $\hat{\tilde{\bm{\tau}}}\in\mathbb{R}^{37}$. Parameter counts range from $0.36$\,M (MLP) to $13.86$\,M (Mamba2).


\section{Per-Joint Torque Prediction}
\label{app:perjoint}

Table~\ref{tab:tau_supp_perjoint} reports per-joint RMSE, MAE, and Pearson correlation for all six baselines on the ForceBody dataset, with right/left joints averaged. We list only the 14 anatomical joints (20 DOFs) with a direct SKEL correspondence (bilateral hip, knee, ankle, MTP, elbow, radioulnar, and wrist), matching the joint set used for $\sigma_\tau$ in Section~\ref{app:uncertainty_stats}. The 6 floating-base pelvis DOFs and the indirect spine, scapula, and shoulder DOFs are excluded because their inverse-dynamics output is either a re-expression of ground reaction (pelvis, already covered by the contact-dynamics task in the main paper), or carries no direct AddB anatomical anchor.

\begin{table*}[!htbp]
\centering
\setlength{\abovecaptionskip}{0pt}
\setlength{\belowcaptionskip}{10pt}
\caption{\textbf{Per-joint torque $\tau$ prediction} in mass-normalized Nm/kg on the ForceBody dataset.}
\label{tab:tau_supp_perjoint}
\scriptsize
\setlength{\tabcolsep}{3pt}
\resizebox{\textwidth}{!}{%
\begin{tabular}{l|ccc|ccc|ccc|ccc|ccc|ccc}
\toprule
\multirow{2}{*}{\textbf{Joint}} & \multicolumn{3}{c|}{\textbf{MLP}} & \multicolumn{3}{c|}{\textbf{Conv1D}} & \multicolumn{3}{c|}{\textbf{LSTM}} & \multicolumn{3}{c|}{\textbf{GRU}} & \multicolumn{3}{c|}{\textbf{Mamba2}} & \multicolumn{3}{c}{\textbf{Transformer}} \\
\cmidrule(lr){2-4} \cmidrule(lr){5-7} \cmidrule(lr){8-10} \cmidrule(lr){11-13} \cmidrule(lr){14-16} \cmidrule(lr){17-19}
 & R$\downarrow$ & M$\downarrow$ & P$\uparrow$ & R$\downarrow$ & M$\downarrow$ & P$\uparrow$ & R$\downarrow$ & M$\downarrow$ & P$\uparrow$ & R$\downarrow$ & M$\downarrow$ & P$\uparrow$ & R$\downarrow$ & M$\downarrow$ & P$\uparrow$ & R$\downarrow$ & M$\downarrow$ & P$\uparrow$ \\
\midrule
\multicolumn{19}{c}{\textit{Upper body}} \\
\midrule
elbow & 0.022 & 0.013 & \red{0.842} & \red{0.020} & \red{0.010} & \red{0.867} & \red{0.020} & \red{0.010} & \red{0.867} & \red{0.019} & \red{0.010} & \red{0.871} & \red{0.019} & \red{0.010} & \red{0.875} & \red{0.023} & \red{0.014} & \red{0.830} \\
pro\_sup & \red{0.003} & 0.001 & \red{0.253} & \red{0.003} & 0.001 & \red{0.371} & \red{0.003} & 0.001 & \red{0.342} & \red{0.003} & 0.001 & \red{0.354} & \red{0.003} & 0.001 & \red{0.385} & \red{0.003} & 0.001 & \red{0.315} \\
wrist\_flex & 0.003 & 0.002 & \red{0.792} & \red{0.003} & \red{0.001} & \red{0.814} & \red{0.003} & \red{0.001} & \red{0.809} & \red{0.003} & \red{0.001} & \red{0.816} & \red{0.002} & \red{0.001} & \red{0.826} & 0.003 & 0.002 & \red{0.767} \\
wrist\_dev & 0.003 & 0.002 & \red{0.689} & 0.003 & 0.002 & \red{0.746} & 0.003 & 0.002 & \red{0.748} & \red{0.003} & \red{0.001} & \red{0.764} & \red{0.003} & \red{0.001} & \red{0.772} & 0.003 & 0.002 & \red{0.692} \\
\midrule
\textbf{Upper mean} & 0.008 & 0.004 & \red{0.644} & 0.007 & 0.004 & \red{0.700} & \red{0.007} & \red{0.004} & \red{0.691} & \red{0.007} & \red{0.003} & \red{0.701} & \red{0.007} & \red{0.003} & \red{0.714} & \red{0.008} & \red{0.005} & \red{0.651} \\
\midrule
\multicolumn{19}{c}{\textit{Lower body}} \\
\midrule
hip\_flex & \red{0.367} & \red{0.264} & \red{0.858} & \red{0.327} & \red{0.230} & \red{0.886} & \red{0.334} & \red{0.235} & \red{0.882} & \red{0.321} & \red{0.226} & \red{0.891} & \red{0.319} & \red{0.223} & \red{0.893} & \red{0.440} & \red{0.323} & \red{0.844} \\
hip\_add & \red{0.377} & \red{0.258} & \red{0.760} & \red{0.316} & \red{0.206} & \red{0.833} & \red{0.331} & \red{0.215} & \red{0.818} & \red{0.314} & \red{0.205} & \red{0.836} & \red{0.304} & \red{0.193} & \red{0.847} & \red{0.385} & \red{0.265} & \red{0.768} \\
hip\_rot & \red{0.141} & \red{0.093} & \red{0.639} & \red{0.127} & \red{0.079} & \red{0.707} & \red{0.133} & \red{0.084} & \red{0.679} & \red{0.126} & \red{0.078} & \red{0.715} & \red{0.125} & \red{0.076} & \red{0.717} & 0.138 & \red{0.090} & \red{0.665} \\
\red{knee} & \red{0.265} & \red{0.188} & \red{0.888} & \red{0.241} & \red{0.168} & \red{0.905} & \red{0.246} & \red{0.172} & \red{0.903} & \red{0.241} & \red{0.166} & \red{0.908} & \red{0.237} & \red{0.160} & \red{0.909} & \red{0.279} & \red{0.200} & \red{0.883} \\
\red{ankle} & \red{0.298} & \red{0.207} & \red{0.915} & \red{0.247} & \red{0.163} & \red{0.942} & \red{0.242} & \red{0.163} & \red{0.945} & 0.243 & \red{0.159} & \red{0.944} & \red{0.238} & \red{0.153} & \red{0.947} & \red{0.272} & \red{0.175} & \red{0.938} \\
mtp & 0.001 & 0.001 & \red{0.938} & 0.001 & 0.001 & \red{0.950} & \red{0.001} & \red{0.000} & \red{0.959} & \red{0.001} & \red{0.000} & \red{0.955} & \red{0.001} & \red{0.000} & \red{0.956} & 0.001 & 0.001 & \red{0.901} \\
\midrule
\textbf{Lower mean} & \red{0.241} & \red{0.168} & \red{0.833} & \red{0.210} & \red{0.141} & \red{0.870} & \red{0.214} & \red{0.145} & \red{0.864} & \red{0.207} & \red{0.139} & \red{0.875} & \red{0.204} & \red{0.134} & \red{0.878} & \red{0.253} & \red{0.176} & \red{0.833} \\
\midrule
\textbf{All joints} & \red{0.148} & \red{0.103} & \red{0.757} & \red{0.129} & \red{0.086} & \red{0.802} & \red{0.131} & \red{0.088} & \red{0.795} & \red{0.127} & \red{0.085} & \red{0.805} & \red{0.125} & \red{0.082} & \red{0.813} & \red{0.155} & \red{0.107} & \red{0.760} \\
\bottomrule
\end{tabular}%
}
\\[2pt]
\begin{flushleft}\footnotesize R: RMSE (Nm/kg)$\downarrow$ \quad M: MAE (Nm/kg)$\downarrow$ \quad P: PCC$\uparrow$\end{flushleft}
\end{table*}

\section{Input Feature Ablation}\label{app:input_ablation}

Here we ablate the parametric body channel by training a Transformer on poses alone and on poses paired with $\bm{\beta}$, mass, and height, all under an identical training protocol. Table~\ref{tab:supp_input_ablation} reports the result. \red{Over three seeds, adding} the parametric body inputs \red{does not reduce} the \red{error. The} All joints RMSE \red{is} \red{$0.159$} \red{with} \red{pose} \red{only} and \red{$0.163$} with the parametric body \red{inputs,} \red{and the PCC is $0.675$ and $0.666$. Because the torque labels are already normalized by body mass, the body inputs add} little \red{information} \red{beyond} \red{the} \red{pose.} \red{Adding pose and root velocities and accelerations, the full input of Tab.~\ref{tab:tau_aggregate}, raises the PCC to $0.760$.}

\begin{table}[h]
\centering
\small
\caption{Input feature ablation on the Transformer baseline. \red{Body} inputs (SKEL $\bm{\beta}$, mass, height) \red{do} \red{not} \red{reduce} \red{the error, while motion} inputs \red{(velocities} \red{and} \red{accelerations)} \red{raise} \red{the} PCC.}
\label{tab:supp_input_ablation}
\setlength{\tabcolsep}{6pt}
\resizebox{\textwidth}{!}{%
\begin{tabular}{l|ccc|ccc|ccc|ccc}
\toprule
\multirow{2}{*}{\textbf{Condition}} & \multicolumn{3}{c|}{\textbf{Inputs}} & \multicolumn{3}{c|}{\textbf{Upper mean}} & \multicolumn{3}{c|}{\textbf{Lower mean}} & \multicolumn{3}{c}{\textbf{All joints}} \\
\cmidrule(lr){2-4} \cmidrule(lr){5-7} \cmidrule(lr){8-10} \cmidrule(lr){11-13}
 & poses & shape & \red{motion}
 & R$\downarrow$ & M$\downarrow$ & P$\uparrow$
 & R$\downarrow$ & M$\downarrow$ & P$\uparrow$
 & R$\downarrow$ & M$\downarrow$ & P$\uparrow$ \\
\midrule
\textbf{(A) poses only}
 & \checkmark & &
 & \red{0.008} & \red{0.005} & \red{0.568}
 & \red{0.260} & \red{0.174} & \red{0.746}
 & \red{0.159} & \red{0.106} & \red{0.675} \\
\textbf{(B) +parametric body}
 & \checkmark & \checkmark &
 & \red{0.009} & \red{0.005} & \red{0.546}
 & \red{0.267} & \red{0.178} & \red{0.745}
 & \red{0.163} & \red{0.109} & \red{0.666} \\
\red{\textbf{(C) +motion (full)}}
 & \red{\checkmark} & \red{\checkmark} & \red{\checkmark}
 & \red{0.008} & \red{0.005} & \red{0.651}
 & \red{0.253} & \red{0.176} & \red{0.833}
 & \red{0.155} & \red{0.107} & \red{0.760} \\
\bottomrule
\end{tabular}%
}
\\[2pt]
\begin{flushleft}\footnotesize R: RMSE (Nm/kg) \quad M: MAE (Nm/kg) \quad P: PCC.\end{flushleft}
\end{table}


\newpage
\providecommand{\answerYes}[1][]{\textcolor{blue}{[Yes] #1}}
\providecommand{\answerNo}[1][]{\textcolor{orange}{[No] #1}}
\providecommand{\answerNA}[1][]{\textcolor{gray}{[N/A] #1}}

\section*{NeurIPS Paper Checklist}

\begin{enumerate}

\item {\bf Claims}
 \item[] Question: Do the main claims made in the abstract and introduction accurately reflect the paper's contributions and scope?
 \item[] Answer: \answerYes{}
 \item[] Justification: The abstract and introduction state three contributions and we deliver each one in the paper. The force-paired parametric body dataset is described in Section 3 (10,386 trials, 26.9 hours, 140 subjects). The per-sample uncertainty annotation is described in Section 3.2 with the Monte Carlo procedure. The reference baselines are reported in Section 4 (GRF prediction against the AddBiomechanics baselines and joint torque prediction with six architectures, plus the Transformer with per-sample uncertainty weighting that yields \red{about 5\%} MAE reduction).

\item {\bf Limitations}
 \item[] Question: Does the paper discuss the limitations of the work performed by the authors?
 \item[] Answer: \answerYes{}
 \item[] Justification: Section 5 notes the representational capacity limitation of SKEL, that $\sigma_\tau$ covers joint-angle noise only, and that the data are dominated by walking and running. Appendix A documents the source and activity imbalance. The Croissant Responsible AI metadata accompanying the dataset additionally documents data limitations, biases, and the isotropic per-DOF noise approximation in detail.

\item {\bf Theory assumptions and proofs}
 \item[] Question: For each theoretical result, does the paper provide the full set of assumptions and a complete (and correct) proof?
 \item[] Answer: \answerNA{}
 \item[] Justification: ForceBody is a dataset and benchmark contribution. The paper does not introduce theoretical results.

\item {\bf Experimental result reproducibility}
 \item[] Question: Does the paper fully disclose all the information needed to reproduce the main experimental results of the paper to the extent that it affects the main claims and/or conclusions of the paper (regardless of whether the code and data are provided or not)?
 \item[] Answer: \answerYes{}
 \item[] Justification: Section 3 specifies the SKEL fitting pipeline (two stages, loss terms, regularizers, source sampling rates, and the resampling to 100\,Hz). Section 3.2 specifies the Monte Carlo procedure (input-side Gaussian noise with $\sigma_q=1.6^\circ$ at the 20 direct DOFs, $N=1000$ samples per frame, GRF held fixed). Section 4 specifies architectures, input features, training schedule, and evaluation protocol. Full hyperparameters and the input ablation are in Appendices B and I.

\item {\bf Open access to data and code}
 \item[] Question: Does the paper provide open access to the data and code, with sufficient instructions to faithfully reproduce the main experimental results, as described in supplemental material?
 \item[] Answer: \answerYes{}
 \item[] Justification: The dataset is hosted publicly on Hugging Face (\url{https://huggingface.co/datasets/kwonjoon/ForceBody}) \red{under CC BY-NC 4.0} with a README documenting the per-trial schema, manifest columns, \red{and} splits. The Croissant metadata file describes the distribution and record sets. \red{The GitHub repository linked in the abstract (\url{https://github.com/DominoAI-Lab/ForceBody-NeurIPS-ED-2026}) releases a data loader, the benchmark scorer, and the code to train and evaluate all baselines, including the uncertainty-weighted finetuning.}

\item {\bf Experimental setting/details}
 \item[] Question: Does the paper specify all the training and test details (e.g., data splits, hyperparameters, how they were chosen, type of optimizer) necessary to understand the results?
 \item[] Answer: \answerYes{}
 \item[] Justification: Section 4 and Appendices G and I specify the train/test split (subject-level, 121 train / 19 test subjects), input features (SKEL pose, optionally shape and motion), output targets (mass-normalized joint torque on the 14 directly matched joints (20 DOFs)), evaluation metrics (RMSE, MAE, Pearson correlation), and all six architectures with their dimensions. \red{The uncertainty-weighted finetuning assigns each training window a loss weight computed from its mean $\sigma_\tau$.}

\item {\bf Experiment statistical significance}
 \item[] Question: Does the paper report error bars suitably and correctly defined or other appropriate information about the statistical significance of the experiments?
 \item[] Answer: \answerYes{}
 \item[] Justification: For fitting quality, we report the full distribution (Mean, Median, P5, P95) of per-trial MPJPE, PA-MPJPE, velocity error, acceleration error, foot sliding, and ground penetration across all 10,386 released trials (Appendix F, Tables 12 and 13), together with acceleration Pearson correlations (\red{0.97}--1.00, Fig.~5). For uncertainty quality, we report the cross-trial median and interquartile range (P25, P75) of $\sigma_\tau$ over the 14 direct joints (Table 2a, Appendix D Table 8), per-activity and per-source breakdowns (Tables 9, 10), the joint-by-activity cross-tabulation (Table 11), Spearman correlations between $\sigma_\tau/|\tau|$ and joint-velocity and torque RMS ($\rho=-0.85$ and $-0.81$, Section 3.2), the $\tau_\text{SKEL}$ RMSE and $\pm\sigma_\tau$ coverage (Section 3.3), and Monte Carlo convergence statistics (standard error of $\hat{\sigma}_\tau$ shrinks from 7.1\% at $N=100$ to 2.2\% at $N=1{,}000$). For the GRF and torque-prediction baselines (Tables 3, 4, 5), we report RMSE, MAE, and Pearson correlation on the held-out test subjects \red{as the mean and standard deviation over three training seeds}.

\item {\bf Experiments compute resources}
 \item[] Question: For each experiment, does the paper provide sufficient information on the computer resources (type of compute workers, memory, time of execution) needed to reproduce the experiments?
 \item[] Answer: \answerYes{}
 \item[] Justification: Training of each baseline takes a few hours on a single NVIDIA L40S GPU (Appendix G). The Monte Carlo annotation pipeline runs on CPU and takes approximately one hour per trial at $N=1000$ samples. The full release was produced over a multi-day batch.

\item {\bf Code of ethics}
 \item[] Question: Does the research conducted in the paper conform, in every respect, with the NeurIPS Code of Ethics \url{https://neurips.cc/public/EthicsGuidelines}?
 \item[] Answer: \answerYes{}
 \item[] Justification: The work uses publicly released, de-identified motion-capture data from the AddBiomechanics dataset under its CC BY 4.0 license. No new human-subject data is collected. The released annotations (SKEL parameters, Monte Carlo torque uncertainty) are computational derivations and do not introduce additional risk to subjects.

\item {\bf Broader impacts}
 \item[] Question: Does the paper discuss both potential positive societal impacts and negative societal impacts of the work performed?
 \item[] Answer: \answerYes{}
 \item[] Justification: Positive impacts (biomechanical assessment, exoskeleton and prosthetic design, sports analytics, physics-aware animation) and the misuse-risk analysis are documented in the Croissant Responsible AI metadata fields \texttt{rai:dataSocialImpact}, \texttt{rai:dataLimitations}, and \texttt{rai:personalSensitiveInformation} accompanying the release. Section 5 of the paper notes the limitations of the release.

\item {\bf Safeguards}
 \item[] Question: Does the paper describe safeguards that have been put in place for responsible release of data or models that have a high risk for misuse (e.g., pre-trained language models, image generators, or scraped datasets)?
 \item[] Answer: \answerYes{}
 \item[] Justification: ForceBody re-expresses already de-identified AddBiomechanics data and introduces no new identifying information. The released file paths contain no user, institution, or system identifiers. \red{The release is distributed under CC BY-NC 4.0, which does not permit commercial use.} Use of the SKEL parameters for mesh instantiation also requires the upstream SKEL model under its non-commercial research license.

\item {\bf Licenses for existing assets}
 \item[] Question: Are the creators or original owners of assets (e.g., code, data, models), used in the paper, properly credited and are the license and terms of use explicitly mentioned and properly respected?
 \item[] Answer: \answerYes{}
 \item[] Justification: The upstream AddBiomechanics dataset (CC BY 4.0) and the source studies aggregated by it (Carter2023, Moore2015, Hamner2013, Falisse2017, Fregly2012, Han2023, Li2021, Tiziana2019, Uhlrich2023, vanderZee2022) are cited in Section 3, Appendix A, and the references. The SKEL body model is cited and its non-commercial research license is noted in the dataset README. \red{The release is distributed under CC BY-NC 4.0, because the SKEL-derived fields follow the non-commercial SKEL license. Code is released under the MIT license.}

\item {\bf New assets}
 \item[] Question: Are new assets introduced in the paper well documented and is the documentation provided alongside the assets?
 \item[] Answer: \answerYes{}
 \item[] Justification: The release ships a README documenting the full per-trial schema, a per-trial manifest, a per-subject summary, a changelog of the release versions, a benchmark card with a scorer, and schema-aligned Croissant metadata that conforms to Croissant 1.1 and the Responsible AI 1.0 extension. \red{The dataset is released under CC BY-NC 4.0 and the code under the MIT license.}

\item {\bf Crowdsourcing and research with human subjects}
 \item[] Question: For crowdsourcing experiments and research with human subjects, does the paper include the full text of instructions given to participants and screenshots, if applicable, as well as details about compensation (if any)?
 \item[] Answer: \answerNA{}
 \item[] Justification: No crowdsourcing was used. ForceBody re-expresses existing AddBiomechanics motion records. No new human-subject data was collected by the authors and no human annotators were involved in producing the released annotations.

\item {\bf Institutional review board (IRB) approvals or equivalent for research with human subjects}
 \item[] Question: Does the paper describe potential risks incurred by study participants, whether such risks were disclosed to the subjects, and whether Institutional Review Board (IRB) approvals (or an equivalent approval/review based on the requirements of your country or institution) were obtained?
 \item[] Answer: \answerNA{}
 \item[] Justification: No new human-subject research was conducted by the authors. The upstream motion records were collected under each source study's own IRB-approved protocol as documented by the AddBiomechanics dataset.

\item {\bf Declaration of LLM usage}
 \item[] Question: Does the paper describe the usage of LLMs if it is an important, original, or non-standard component of the core methods in this research? Note that if the LLM is used only for writing, editing, or formatting purposes and does not impact the core methodology, scientific rigor, or originality of the research, declaration is not required.
 \item[] Answer: \answerNA{}
 \item[] Justification: LLMs are not part of any core method in this paper.

\end{enumerate}

\end{document}


\maketitle



\section{Per-Dataset Composition}
\label{app:dataset_composition}

ForceBody aggregates motion trials from ten public motion-capture datasets included in AddBiomechanics~\cite{werling2024addbiomechanics}. After applying the quality-based filter (Suppl.~\ref{app:p80_filter}), the released dataset spans 140 subjects, 10,389 trials, and approximately 26.9 hours of recorded motion at 100\,Hz. Table~\ref{tab:dataset_composition} summarizes the per-source composition: trial and subject counts, the male/female/unreported breakdown, mean$\pm$std subject height and mass, and the dominant activity type.

The dataset is dominated by the Carter2023~\cite{carter2024full} running-and-walking corpus ($67\%$ of trials), with the remaining trials covering walking, stair climbing and jumping. Demographics span a representative range of adult anthropometry (height $1.73\pm0.09$\,m, mass $68.8\pm11.5$\,kg) but skew toward male subjects ($86/140 = 61\%$); ten subjects from vanderZee2022~\cite{van2022biomechanics} have unreported sex. We retain all subjects regardless of sex availability; the SKEL body model is fit using a male template when sex is unreported, following AddBiomechanics' default convention.

\begin{table}[h]
\centering
\small
\caption{Per-dataset composition of \textbf{ForceBody}. Subjects refers to unique individuals; each subject may contribute multiple recording sessions. M/F denotes male/female; U = unreported.}
\label{tab:dataset_composition}
\begin{tabular}{lrrcccc}
\toprule
Source & Trials & Subj. & M/F/U & Height (m) & Mass (kg) & Activity \\
\midrule
Carter2023~\cite{carter2024full}
  & $7{,}006$ & $50$ & 25/25/0 & $1.71{\pm}0.06$ & $65.5{\pm}9.9$  & Walk / run \\
vanderZee2022~\cite{van2022biomechanics}
  & $1{,}089$ & $10$ & 0/0/10  & $1.74{\pm}0.10$ & $71.7{\pm}13.9$ & Walking \\
Moore2015~\cite{moore2015elaborate}
  & $1{,}019$ & $12$ & 8/4/0   & $1.72{\pm}0.08$ & $69.7{\pm}10.9$ & Walking \\
Han2023~\cite{han2023groundlink}
  & $243$     & $7$  & 3/4/0   & $1.71{\pm}0.12$ & $67.1{\pm}11.3$ & Jump / cut \\
Uhlrich2023~\cite{uhlrich2023opencap}
  & $151$     & $9$  & 3/6/0   & $1.72{\pm}0.09$ & $68.2{\pm}11.3$ & Walking \\
Fregly2012~\cite{fregly2012grand}
  & $78$      & $4$  & 3/1/0   & $1.69{\pm}0.03$ & $72.2{\pm}6.3$  & Walking \\
Li2021~\cite{li2021well}
  & $65$      & $1$  & 1/0/0   & $1.83$          & $85.7$          & Stairs \\
Falisse2017~\cite{falisse2016emg}
  & $64$      & $9$  & 6/3/0   & $1.76{\pm}0.04$ & $71.3{\pm}5.5$  & Sit-to-stand / walk \\
Hammer2013~\cite{hamner2013muscle}
  & $40$      & $10$ & 10/0/0  & $1.74{\pm}0.05$ & $71.6{\pm}6.1$  & Running \\
Others
  & $634$     & $28$ & 27/1/0  & $1.75{\pm}0.11$ & $69.1{\pm}13.4$ & Mixed \\
\midrule
Total
  & $10{,}389$ & $140$ & 86/44/10 & $1.73{\pm}0.09$ & $68.8{\pm}11.5$ & \\
\bottomrule
\end{tabular}
\end{table}


\section{Fitting Pipeline}
\label{supp:fitting_details}

\begin{algorithm}[h]
\caption{SKEL fitting (per-trial).}
\label{alg:supp_fitting}
\begin{algorithmic}[1]
\Require AddB joint positions $\mathbf{J}^{\text{GT}}_{1:T}$, marker positions $\mathbf{p}^{\text{obs}}_{1:T}$, reference angles $\bm{\theta}^{\text{ref}}_{1:T}$ with mask $\mathbf{m}$
\Ensure SKEL shape $\bm{\beta}$ and per-frame pose/translation $\{\bm{\theta}_t, \mathbf{t}_t\}_{t=1}^{T}$
\\\hrulefill
\State \textbf{Stage 1 — Shape estimation}
\State $\bar{\bm{\ell}}^{\text{GT}} \gets \text{frame-mean of AddB bone lengths}$
\State $\bm{\beta} \gets \displaystyle\arg\min_{\bm{\beta}} \sum_i w_i\,\bigl(\ell_i^{\text{SKEL}}(\bm{\beta}) - \bar{\ell}_i^{\text{GT}}\bigr)^2$ \Comment{Adam, $200$ iters}
\State Freeze $\bm{\beta}$
\\\hrulefill
\State \textbf{Stage 2 — Pose optimization}
\State Initialize $\{\bm{\theta}_t, \mathbf{t}_t\}$ from pelvis-yaw candidates $\{0^\circ,90^\circ,180^\circ,270^\circ\}$, pick the best
\For{$k = 1, \dots, K_\text{iter}$}
    \State $\mathcal{L}_{\text{pos}} \gets \lambda_\text{j}\mathcal{L}_\text{joint} + \lambda_\text{b}\mathcal{L}_\text{bone} + \lambda_\text{w}\mathcal{L}_\text{width} + \lambda_\text{f}\mathcal{L}_\text{foot} + \lambda_\text{m}\mathcal{L}_\text{marker}$
    \State $\mathcal{L}_{\text{reg}} \gets \lambda_\text{p}\|\bm{\theta}\|^2 + \lambda_\text{s}\|\bm{\theta}_\text{spine}\|^2 + \lambda_\text{sc}\|\bm{\theta}_\text{scap}\|^2 + \lambda_\text{h}\|\bm{\theta}_\text{hum}\|^2$
    \State $\mathcal{L}_{\text{temp}} \gets \dfrac{\lambda_\text{v}}{T}\sum_t\|\dot{\bm{\theta}}_t\|^2 + \dfrac{\lambda_\text{a}}{T}\sum_t\|\ddot{\bm{\theta}}_t\|^2$
    \State $\alpha(k) \gets \max\!\left(0,\ \dfrac{k/K_\text{iter} - 0.3}{0.7}\right)$ \Comment{q-match ramp after $30\%$ warm-up}
    \State $\mathcal{L}_{\text{q-match}} \gets \alpha(k)\cdot\dfrac{1}{T}\sum_t \bigl\|(\bm{\theta}_t - \bm{\theta}^{\text{ref}}_t)\odot\mathbf{m}\bigr\|^2$
    \State $\mathcal{L} \gets \mathcal{L}_{\text{pos}} + \mathcal{L}_{\text{reg}} + \mathcal{L}_{\text{temp}} + \lambda_\text{q}\mathcal{L}_{\text{q-match}}$
    \State $\{\bm{\theta}_t, \mathbf{t}_t\} \gets \text{Adam step on } \nabla\mathcal{L}$
    \State \textbf{break} if early-stop patience exceeded
\EndFor
\State Marker-only fine-tuning ($200$ iters); then foot-only fine-tuning ($400$ iters)
\State \Return $\bm{\beta},\, \{\bm{\theta}_t, \mathbf{t}_t\}_{t=1}^{T}$
\end{algorithmic}
\end{algorithm}

\paragraph{Preprocessing.} Source datasets span native sampling rates of $61$--$251$\,Hz. We apply a 4th-order zero-lag Butterworth lowpass filter (\texttt{filtfilt}, cutoff at $90\%$ Nyquist) and resample to $100$\,Hz via linear interpolation, following the convention of GaitDynamics~\cite{tan2026gaitdynamics} and AddBiomechanics~\cite{werling2024addbiomechanics}. Trials with native rates below $100$\,Hz are kept at their original sampling rate without upsampling. We use kinematics from the inverse-kinematics pass only, excluding the force-based kinematics refinement of AddBiomechanics, to prevent force labels from leaking into kinematic inputs~\cite{tan2026gaitdynamics}.

\paragraph{Two-stage fitting.} For each trial we recover SKEL parameters $(\bm{\beta}, \{\bm{\theta}_t, \mathbf{t}_t\}_{t=1}^T)$ in two sequential stages (Algorithm~\ref{alg:supp_fitting}). \textbf{Stage 1} estimates the per-trial shape parameters $\bm{\beta}\in\mathbb{R}^{10}$ once per trial by matching SKEL bone lengths to AddB target lengths over $10$--$12$ reliable bone pairs (femur, tibia, foot, forearm). The target lengths are averaged across all frames in the trial to provide a single, frame-stable target. The shin bone pair is weighted $10\times$ because SKEL's shape space couples thigh and shin dimensions; without the upweighting, the optimizer trades shin error for thigh error, producing knee-level fitting artifacts. \textbf{Stage 2} fixes $\bm{\beta}$ and optimizes the per-frame pose $\bm{\theta}_t\in\mathbb{R}^{46}$ and translation $\mathbf{t}_t\in\mathbb{R}^{3}$ under the composite loss
\begin{equation}
\mathcal{L} \;=\; \mathcal{L}_{\text{pos}} \;+\; \mathcal{L}_{\text{reg}} \;+\; \mathcal{L}_{\text{temp}} \;+\; \lambda_\text{q}\,\mathcal{L}_{\text{q-match}}.
\label{eq:supp_pose_loss}
\end{equation}

\paragraph{Positional loss.} The positional term $\mathcal{L}_{\text{pos}}$ aggregates five sub-terms:
\begin{equation}
\mathcal{L}_{\text{pos}} \;=\; \lambda_\text{j}\,\mathcal{L}_\text{joint} \;+\; \lambda_\text{b}\,\mathcal{L}_\text{bone} \;+\; \lambda_\text{w}\,\mathcal{L}_\text{width} \;+\; \lambda_\text{f}\,\mathcal{L}_\text{foot} \;+\; \lambda_\text{m}\,\mathcal{L}_\text{marker},
\end{equation}
where $\mathcal{L}_\text{joint}$ is a per-DOF-weighted Huber loss ($\delta{=}50$\,mm) over the 15 directly matched joints, $\mathcal{L}_\text{bone}$ penalizes squared bone-length deviation across 12 pairs, $\mathcal{L}_\text{width}$ aligns shoulder width, $\mathcal{L}_\text{foot}$ enforces vertical alignment of foot joints (critical for downstream GRF consistency), and $\mathcal{L}_\text{marker}$ matches BSM mesh vertices to the observed optical markers.

\paragraph{Regularization and temporal smoothness.} The regularizer $\mathcal{L}_{\text{reg}}$ penalizes unnatural rotations in under-constrained regions where AddB provides no direct anatomical anchor:
\begin{equation}
\mathcal{L}_{\text{reg}} \;=\; \lambda_\text{p}\,\|\bm{\theta}\|^2 + \lambda_\text{s}\,\|\bm{\theta}_\text{spine}\|^2 + \lambda_\text{sc}\,\|\bm{\theta}_\text{scap}\|^2 + \lambda_\text{h}\,\|\bm{\theta}_\text{hum}\|^2.
\end{equation}
The temporal term $\mathcal{L}_{\text{temp}}$ enforces smoothness via finite differences,
\begin{equation}
\mathcal{L}_{\text{temp}} \;=\; \frac{\lambda_\text{v}}{T}\sum_t \|\dot{\bm{\theta}}_t\|^2 + \frac{\lambda_\text{a}}{T}\sum_t \|\ddot{\bm{\theta}}_t\|^2.
\end{equation}

\paragraph{DOF-matching prior.} The optional pose-matching term $\mathcal{L}_\text{q-match}$ softly anchors SKEL DOFs to a per-frame reference pose $\bm{\theta}^{\text{ref}}_t$, which is obtained either by (i) directly estimating lower-body SKEL DOFs from rotation matrices computed against the AddB skeleton, or (ii) by mapping AddB Rajagopal joint angles to SKEL DOF indices via fixed scale factors. The term is gated by a per-DOF mask $\mathbf{m}$ that selects only the DOFs for which a reference is available, and is linearly ramped up from $0$ after a $30\%$ warm-up:
\begin{equation}
\mathcal{L}_\text{q-match} \;=\; \alpha(k) \cdot \frac{1}{T}\sum_t \bigl\|(\bm{\theta}_t - \bm{\theta}^{\text{ref}}_t)\odot\mathbf{m}\bigr\|^2,
\end{equation}
where $\alpha(k)$ ramps linearly from $0$ to $1$ over the post-warm-up iterations. We set $\lambda_\text{q}{=}0.1$, so this term contributes a small fraction of the total loss and primarily serves to stabilize lower-body and mapped DOFs where the AddB reference is available; under-constrained DOFs without an AddB counterpart (spine, scapula, humerus) are instead regularized by $\mathcal{L}_\text{reg}$.

\paragraph{Hyperparameters.} Both stages use Adam with cosine-annealed learning rate $10^{-2}$. Stage 1 runs for $200$ iterations (extended to $1{,}000$ for outlier subjects). Stage 2 runs for $K_\text{iter}{=}500$ iterations on standard trials and $1{,}000$--$2{,}000$ on outlier trials, with early stopping at patience $80$. The Huber loss uses $\delta{=}50$\,mm. To break pelvis-orientation ambiguity, we initialize the pose at four pelvis yaw candidates ($0^\circ, 90^\circ, 180^\circ, 270^\circ$) and keep the lowest-loss initialization. After the main optimization, we run a marker-only fine-tuning ($200$ iterations) followed by a foot-only fine-tuning ($400$ iterations) to clean up residual marker drift and foot-ground contact. The DOF-matching prior is activated only after a $30\%$ warm-up. Loss weights are listed in Table~\ref{tab:supp_loss_weights}.

\begin{table}[h]
\centering
\small
\caption{Stage 2 loss weights.}
\label{tab:supp_loss_weights}
\begin{tabular}{lcc}
\toprule
Term & Symbol & Weight \\
\midrule
Joint position    & $\lambda_\text{j}$  & $1.0$  \\
Bone length       & $\lambda_\text{b}$  & $1.0$  \\
Shoulder width    & $\lambda_\text{w}$  & $2.0$  \\
Foot height       & $\lambda_\text{f}$  & $15.0$ \\
Marker            & $\lambda_\text{m}$  & $10.0$ \\
Pose reg.         & $\lambda_\text{p}$  & $0.01$ \\
Spine reg.        & $\lambda_\text{s}$  & $0.3$  \\
Scapula reg.      & $\lambda_\text{sc}$ & $0.05$ \\
Humerus reg.      & $\lambda_\text{h}$  & $0.05$ \\
Velocity          & $\lambda_\text{v}$  & $0.01$ \\
Acceleration      & $\lambda_\text{a}$  & $0.1$  \\
DOF match         & $\lambda_\text{q}$  & $0.1$  \\
\bottomrule
\end{tabular}
\end{table}


\section{Pseudo-code for per-frame torque uncertainty}
\label{app:mc_pseudocode}

For every trial of \textbf{ForceBody}, we compute the per-frame, per-joint torque uncertainty $\sigma_{\tau}$ by Monte Carlo sampling through the same OpenSim inverse-dynamics (ID) solver that produced the ground-truth torques. We draw $N{=}1000$ band-limited Gaussian perturbations of the AddB joint angles using an FFT-based generator with RMS $\sigma_q{=}1.6^{\circ}$ and a $6$\,Hz lowpass cutoff. Noise is injected only on the 20 DOFs spanning the 14 major joints with a direct SKEL correspondence (bilateral hip, knee, ankle, MTP, elbow, radioulnar, wrist). We add the position noise to the released GT pose and propagate it to velocity and acceleration by forward finite differencing, leaving the GT $\dot{\bm{q}}$ and $\ddot{\bm{q}}$ as the baseline so that the resulting torque variance is attributable to $\sigma_q$ alone. We then run AddB inverse dynamics on the perturbed kinematics with the measured GRF held fixed, and aggregate the per-DOF mean $\mu_{\tau}$ and standard deviation $\sigma_{\tau}$ across the $N$ samples. The procedure is summarized in Algorithm~\ref{alg:mc_validation}.

\begin{algorithm}[t]
\caption{Per-frame torque uncertainty via Monte Carlo sampling.}
\label{alg:mc_validation}
\begin{algorithmic}[1]
\Require AddB kinematics $\bm{q}_{1:T},\, \dot{\bm{q}}_{1:T},\, \ddot{\bm{q}}_{1:T}$;
         measured GRF $\bm{\lambda}_{1:T}$;
         direct-DOF mask $\mathcal{M}$;
         pose RMS $\sigma_q{=}1.6^{\circ}$;
         cutoff $f_c{=}6$\,Hz;
         samples $N{=}1000$
\For{$n = 1, \dots, N$}
  \State $\bm{\epsilon}^{(n)} \gets \mathrm{BandLimitedGaussian}(T,\, \sigma_q,\, f_c)$
         \Comment{zero outside $\mathcal{M}$}
  \State $\dot{\bm{\epsilon}}^{(n)},\, \ddot{\bm{\epsilon}}^{(n)} \gets \mathrm{ForwardDiff}(\bm{\epsilon}^{(n)})$
  \State $\tilde{\bm{\tau}}^{(n)} \gets
          \mathrm{ID}\bigl(\bm{q}{+}\bm{\epsilon}^{(n)},\, \dot{\bm{q}}{+}\dot{\bm{\epsilon}}^{(n)},\, \ddot{\bm{q}}{+}\ddot{\bm{\epsilon}}^{(n)},\, \bm{\lambda}\bigr)$
\EndFor
\State $\mu_{\tau,t},\, \sigma_{\tau,t} \gets
        \mathrm{mean/std}\bigl\{\tilde{\bm{\tau}}_t^{(n)}\bigr\}_{n=1}^{N}$
        \Comment{per frame $t$, per DOF}
\State \Return $\{\mu_{\tau,t},\, \sigma_{\tau,t}\}_{t=1}^{T}$
\end{algorithmic}
\end{algorithm}

\begin{figure}[!t]
\centering
\includegraphics[width=\textwidth]{fig/torque_uncertainty_panels_carter.png}
\caption{\textbf{Per-joint torque uncertainty for one representative frame
(extended).} Same trial and frame as main paper, extended to both right and left sides.}
\label{fig:per_joint_sigma_supp}
\end{figure}

\section{Per-Sample Torque Uncertainty: Statistics}
\label{supp:torque_uncertainty}

This section breaks down the released $\bm{\sigma}_{\tau}$ across joints, activities, and source datasets on the $8{,}767$ released trials. All values are mass-normalized (Nm/kg) and reported as the cross-trial median of the per-trial median $\sigma_{\tau}$.

\paragraph{Per joint.} Table~\ref{tab:supp_sigma_per_joint} reports $\bm{\sigma}_{\tau}$ on the 14 direct joints. Hip and knee carry the widest envelopes ($0.47$ and $0.21$\,Nm/kg) due to their long lever arms. The toe joint (MTP) and the distal forearm joints (radioulnar, wrist) carry the tightest ($\sim\!10^{-3}$\,Nm/kg).

\begin{table}[h]
\centering
\small
\caption{Per-joint $\bm{\sigma}_{\tau}$ across the 14 direct joints. Left and right averaged. \textit{Trial-Med} is the cross-trial median of each per-trial median $\sigma_{\tau}$. P25 and P75 are the inter-quartile range.}
\label{tab:supp_sigma_per_joint}
\begin{tabular}{lccc}
\toprule
Joint & Trial-Med [Nm/kg] & P25 & P75 \\
\midrule
Hip          & $0.472$  & $0.452$  & $0.508$  \\
Knee         & $0.215$  & $0.204$  & $0.232$  \\
Ankle        & $0.119$  & $0.103$  & $0.129$  \\
MTP          & $0.001$  & $0.0008$ & $0.0011$ \\
Elbow        & $0.015$  & $0.013$  & $0.017$  \\
Radioulnar   & $0.0004$ & $0.0004$ & $0.0004$ \\
Wrist        & $0.002$  & $0.0016$ & $0.0019$ \\
\bottomrule
\end{tabular}
\end{table}

\paragraph{Per activity.} Table~\ref{tab:supp_sigma_per_activity} reports $\bm{\sigma}_{\tau}$ and the relative envelope $\sigma_{\tau}/|\tau|$ across activity classes. Dynamic activities (running, jumping) carry the smallest relative envelopes (rel-med $0.47$ for running, $0.75$ for jumping). Near-static activities (quiet standing) carry the largest ($1.26$). Trials whose activity label is missing or does not fall into one of the named classes are pooled in \textit{Others}. The absolute envelope $\sigma_{\tau}$ is roughly comparable across activities ($0.013$--$0.019$\,Nm/kg), so the relative trend is driven mostly by the size of the underlying torque signal $|\tau|$ rather than by changes in $\sigma_{\tau}$ itself.

\begin{table}[h]
\centering
\small
\caption{$\bm{\sigma}_{\tau}$ broken down by activity class. \textit{Sigma-Med} is the cross-trial median of the per-trial median $\sigma_{\tau}$ pooled across the 14 direct joints. \textit{Rel-Med} is the cross-trial median of $\sigma_{\tau}/|\tau|$.}
\label{tab:supp_sigma_per_activity}
\begin{tabular}{lcrc}
\toprule
Activity     & Trials & Sigma-Med [Nm/kg] & Rel-Med \\
\midrule
Running        & $4{,}949$ & $0.013$ & $0.47$ \\
Jumping        & $26$      & $0.018$ & $0.75$ \\
Sit-to-stand   & $22$      & $0.019$ & $0.81$ \\
Others         & $870$     & $0.017$ & $0.76$ \\
Walking        & $2{,}708$ & $0.017$ & $0.81$ \\
Stairs         & $4$       & $0.015$ & $1.20$ \\
Quiet standing & $188$     & $0.017$ & $1.26$ \\
\bottomrule
\end{tabular}
\end{table}

\paragraph{Per source.} Table~\ref{tab:supp_sigma_per_source} reports the same statistics broken down by source dataset. Running-dominated sources (Hammer2013, Carter2023) sit at the low end of the relative envelope ($0.38$ and $0.50$). Sources dominated by slow or quasi-static motion (Falisse2017 sit-to-stand, Fregly2012) sit at the high end ($1.26$ and $1.21$).

\begin{table}[h]
\centering
\small
\caption{$\bm{\sigma}_{\tau}$ broken down by source dataset. \textit{Sigma-Med} and \textit{Rel-Med} follow Table~\ref{tab:supp_sigma_per_activity}.}
\label{tab:supp_sigma_per_source}
\begin{tabular}{lcrc}
\toprule
Source        & Trials & Sigma-Med [Nm/kg] & Rel-Med \\
\midrule
Hammer2013    & $28$      & $0.018$ & $0.38$ \\
Carter2023    & $6{,}383$ & $0.014$ & $0.50$ \\
Tiziana2019   & $1$       & $0.016$ & $0.57$ \\
vanderZee2022 & $1{,}021$ & $0.017$ & $0.74$ \\
Uhlrich2023   & $15$      & $0.018$ & $0.84$ \\
Han2023       & $169$     & $0.019$ & $0.84$ \\
Moore2015     & $326$     & $0.016$ & $0.86$ \\
Fregly2012    & $62$      & $0.019$ & $1.21$ \\
Falisse2017   & $22$      & $0.016$ & $1.26$ \\
\bottomrule
\end{tabular}
\end{table}

\paragraph{Joint-by-activity relative uncertainty.} Table~\ref{tab:supp_sigma_joint_activity} cross-tabulates $\sigma_{\tau}/|\tau|$ by joint and activity on the 14 direct joints. Hip is the joint whose relative uncertainty changes the most across activities, dropping to $0.77$ in running and rising to $3.43$ in quiet standing, where hip torque is dominated by quasi-static body-weight support and the kinematic-noise contribution becomes a larger fraction of the signal. Knee shows a similar pattern but on a smaller scale, while distal joints (MTP, radioulnar, wrist) stay closer across activities.

\begin{table}[h]
\centering
\small
\caption{Joint-by-activity median relative uncertainty $\sigma_{\tau}/|\tau|$ on the 14 direct joints.}
\label{tab:supp_sigma_joint_activity}
\setlength{\tabcolsep}{4pt}
\begin{tabular}{lccccccc}
\toprule
Joint      & Run  & Jump & Sit-to-stand & Others & Walk & Stair & Standing \\
\midrule
Hip        & $0.77$ & $2.81$ & $0.95$ & $1.69$ & $1.34$ & $2.13$ & $3.43$ \\
Knee       & $0.40$ & $0.46$ & $0.19$ & $0.69$ & $0.52$ & $0.42$ & $2.17$ \\
Ankle      & $0.24$ & $0.34$ & $0.42$ & $0.30$ & $0.17$ & $1.03$ & $0.38$ \\
MTP        & $0.52$ & $0.71$ & $0.93$ & $0.65$ & $0.68$ & $0.84$ & $0.86$ \\
Elbow      & $0.41$ & $1.39$ & $0.65$ & $1.07$ & $0.93$ & $1.29$ & $0.93$ \\
Radioulnar & $0.56$ & $0.74$ & $0.70$ & $0.68$ & $0.86$ & $1.10$ & $2.17$ \\
Wrist      & $0.48$ & $1.10$ & $0.73$ & $1.10$ & $1.07$ & $1.94$ & $1.33$ \\
\bottomrule
\end{tabular}
\end{table}

\section{Quality-based filtering of \textbf{ForceBody}}
\label{app:p80_filter}

The raw trial data can contain heavy-tailed outlier trials, typically caused by very fast running, sharp cuts, or abnormal body shapes. To ensure data quality, all quantitative results in this paper are computed on a quality-filtered subset of \textbf{ForceBody}. Specifically, we consider the threshold suggested in~\cite{hicks2015my} and define a single hard cutoff. This leads to a subset with root-relative MPJPE is at or below the $80$th percentile of the unfiltered baseline. This filtering improves data quality while preserving all $10$ source datasets and all $7$ activity categories. The resulting subset contains $10{,}389$ trials ($80.0\%$ of $12{,}986$), $9{,}689{,}847$ frames at $100$\,Hz ($\sim 26.9$ hours), and $140$ of the original $143$ unique subjects ($97.9\%$).


\section{Fitting Quality of ForceBody}\label{supp:fitting_quality}

This section provides the full breakdown of fitting quality omitted from
the main paper, evaluated on all $10{,}389$ released trials. Per-joint
MPJPE values are reported on the $15$ joints with a direct
correspondence between the AddB Rajagopal skeleton and the SKEL
kinematic tree, with left and right sides averaged.

\textbf{Aggregate metrics.} We report two complementary metric groups
in Table~\ref{tab:supp_fitting_quality}, top. \textbf{Positional errors}
compare predicted SKEL joint centers to the AddB joint centers. MPJPE
is the mean per-joint Euclidean distance in world coordinates, and
PA-MPJPE applies a per-frame Procrustes alignment that isolates shape
from pose-and-position error. \textbf{Physical-plausibility errors}
measure how realistic the resulting motion is. Per-frame velocity and
acceleration errors (mm/f, mm/f$^2$) capture trajectory smoothness, and
foot sliding and ground penetration (both in mm) capture foot-contact
artifacts. Table~\ref{tab:fitting_quality_full} shows the comprehensive analysis on the fitting quality.

\textbf{Per-joint breakdown.} Table~\ref{tab:supp_fitting_quality},
bottom, reports per-joint MPJPE on the $15$ direct joints. Lower-body
joints (pelvis, hip, knee, ankle, MTP) average $16.8$\,mm. Upper-body
joints (elbow, forearm, wrist) average $27.5$\,mm. The upper-body
errors are larger across all three joints, consistent with the wider
operating range of the arms in everyday motion and the sparser optical
marker coverage of the upper body in the source datasets.

\begin{table}[t]
\centering
\small
\caption{\textbf{Top}: Aggregate fitting quality of \textbf{ForceBody}.
\textbf{Bottom}: per-joint MPJPE on the $15$ directly matched joints,
left and right averaged. Pelv.=Pelvis, Kne.=Knee, Ank.=Ankle,
MTP=Metatarsophalangeal joint, F.a.=Forearm (radioulnar), Wri.=Wrist,
Elb.=Elbow.}
\label{tab:supp_fitting_quality}
\begin{tabular*}{0.85\linewidth}{@{\extracolsep{\fill}} l cc cccc @{}}
\toprule
 & \multicolumn{2}{c}{Positional Err.} & \multicolumn{4}{c}{Physical-plausibility Err.} \\
\cmidrule(lr){2-3} \cmidrule(lr){4-7}
 & MPJPE & PA-MPJPE & Vel.\ err.\ & Acc.\ err.\ & Foot slid.\ & Ground pen.\ \\
 & [mm]  & [mm]     & [mm/f]      & [mm/f$^2$]   & [mm]        & [mm]         \\
\midrule
\textbf{ForceBody} & $21.3$ & $20.3$ & $1.40$ & $1.19$ & $1.08$ & $1.26$ \\
\midrule
\midrule
\end{tabular*}%
\\[-\arrayrulewidth]
\begin{tabular*}{0.85\linewidth}{@{\extracolsep{\fill}} l ccccc ccc @{}}
 & \multicolumn{5}{c}{Lower body (Avg.\ 16.8)} & \multicolumn{3}{c}{Upper body (Avg.\ 27.5)} \\
\cmidrule(lr){2-6} \cmidrule(lr){7-9}
 & MTP & Ank. & Kne. & Hip & Pelv. & Wri. & F.a. & Elb. \\
\midrule
Mean & $16.4$ & $19.0$ & $19.4$ & $15.7$ & $13.4$ & $27.6$ & $26.6$ & $28.4$ \\
\bottomrule
\end{tabular*}
\end{table}

\begin{table}[h]
\centering
\small
\setlength{\tabcolsep}{6pt}
\caption{Full distribution of per-trial fitting-quality metrics across
the quality-filtered subset of \textbf{ForceBody} ($10{,}389$ trials).}
\label{tab:fitting_quality_full}
\begin{tabular}{lrrrr}
\toprule
Metric & Mean & Median & P$5$ & P$95$ \\
\midrule
MPJPE (mm)                    & $21.3$ & $22.0$ & $11.4$ & $31.5$ \\
PA-MPJPE (mm)                 & $20.3$ & $20.8$ & $10.9$ & $30.6$ \\
Velocity error (mm/f)         & $1.40$ & $1.33$ & $0.51$ & $2.74$ \\
Acceleration error (mm/f$^2$) & $1.19$ & $0.76$ & $0.41$ & $4.41$ \\
Foot sliding (mm)             & $1.08$ & $0.00$ & $0.00$ & $7.18$ \\
Ground penetration (mm)       & $1.26$ & $0.16$ & $0.00$ & $8.09$ \\
\bottomrule
\end{tabular}
\end{table}

\section{Joint Torque Training Details}\label{supp:joint_torque_training}

This section provides the training details for the joint torque baselines reported in the main paper.

\paragraph{Targets and masking.}
The supervision target at each frame is the AddB inverse-dynamics torque vector $\bm{\tau}_t \in \mathbb{R}^{37}$ on the Rajagopal skeleton. We restrict the loss to the 15 anatomical joints (20 DOFs) with a direct SKEL correspondence (bilateral hip, knee, ankle, MTP, elbow, radioulnar, and wrist), matching the joint set used for $\sigma_\tau$ in Section~\ref{supp:torque_uncertainty}. A binary DOF mask $\mathbf{m} \in \{0,1\}^{37}$ selects these direct DOFs, with all other DOFs (pelvis floating-base, lumbar, scapula, humerus) masked out. We additionally mask any frame whose GRF coverage is invalid in the AddB labels.

\paragraph{Per-DOF $z$-score normalization.}
Joint torque magnitudes vary by two orders of magnitude across the body, from $\sim\!0.01$\,Nm/kg at the wrist and toe joints to $\sim\!1$\,Nm/kg at the hip. Training a single model on raw torque values is therefore dominated by the high-magnitude lower-body joints. To balance the loss across joints, we normalize each DOF independently by its training-set mean $\mu_d$ and standard deviation $\sigma_d$,
\begin{equation}
\tilde{\tau}_{t,d} \;=\; \frac{\tau_{t,d} - \mu_d}{\sigma_d}, \quad d \in \{1, \dots, 37\},
\end{equation}
so that every DOF contributes comparable magnitude to the loss. The model predicts $\tilde{\tau}$ in normalized space, and we de-normalize at evaluation time to report errors in physical Nm/kg.

\paragraph{Loss.}
The training loss is the masked mean-squared error over the direct DOFs and valid frames in a batch,
\begin{equation}
\mathcal{L}_\tau \;=\; \frac{1}{\sum_{t,d} \mathbf{m}_d \cdot v_t}\sum_{t,d} \mathbf{m}_d \cdot v_t \cdot \bigl(\hat{\tilde{\tau}}_{t,d} - \tilde{\tau}_{t,d}\bigr)^2,
\end{equation}
where $v_t \in \{0, 1\}$ is the per-frame validity flag.

\paragraph{Optimization.}
All baselines are optimized with Adam at learning rate $2{\times}10^{-4}$, batch size $128$, and sequence length $256$ frames ($2.56$\,s at $100$\,Hz). We train each model on a single NVIDIA L40S GPU until the validation loss plateaus.


\section{Per-Joint Torque Prediction}
\label{app:tau_perjoint}

Table~\ref{tab:tau_supp_perjoint} reports per-joint RMSE, MAE, and Pearson correlation for all six baselines on the ForceBody dataset, with right/left joints averaged. We list only the 14 anatomical joints (20 DOFs) with a direct SKEL correspondence (bilateral hip, knee, ankle, MTP, elbow, radioulnar, and wrist), matching the joint set used for $\sigma_\tau$ in Section~\ref{supp:torque_uncertainty}. The 6 floating-base pelvis DOFs and the indirect spine, scapula, and shoulder DOFs are excluded because their inverse-dynamics output is either a re-expression of ground reaction (pelvis, already covered by the contact-dynamics task in the main paper), or carries no direct AddB anatomical anchor.

\begin{table*}[!htbp]
\centering
\setlength{\abovecaptionskip}{0pt}
\setlength{\belowcaptionskip}{10pt}
\caption{\textbf{Per-joint torque $\tau$ prediction} in mass-normalized Nm/kg on the ForceBody dataset.}
\label{tab:tau_supp_perjoint}
\scriptsize
\setlength{\tabcolsep}{3pt}
\resizebox{\textwidth}{!}{%
\begin{tabular}{l|ccc|ccc|ccc|ccc|ccc|ccc}
\toprule
\multirow{2}{*}{\textbf{Joint}} & \multicolumn{3}{c|}{\textbf{MLP}} & \multicolumn{3}{c|}{\textbf{Conv1D}} & \multicolumn{3}{c|}{\textbf{LSTM}} & \multicolumn{3}{c|}{\textbf{GRU}} & \multicolumn{3}{c|}{\textbf{Mamba2}} & \multicolumn{3}{c}{\textbf{Transformer}} \\
\cmidrule(lr){2-4} \cmidrule(lr){5-7} \cmidrule(lr){8-10} \cmidrule(lr){11-13} \cmidrule(lr){14-16} \cmidrule(lr){17-19}
 & R$\downarrow$ & M$\downarrow$ & P$\uparrow$ & R$\downarrow$ & M$\downarrow$ & P$\uparrow$ & R$\downarrow$ & M$\downarrow$ & P$\uparrow$ & R$\downarrow$ & M$\downarrow$ & P$\uparrow$ & R$\downarrow$ & M$\downarrow$ & P$\uparrow$ & R$\downarrow$ & M$\downarrow$ & P$\uparrow$ \\
\midrule
\multicolumn{19}{c}{\textit{Upper body}} \\
\midrule
elbow & 0.022 & 0.012 & 0.757 & 0.022 & 0.012 & 0.747 & 0.022 & 0.013 & 0.750 & 0.022 & 0.013 & 0.751 & 0.021 & 0.011 & 0.776 & 0.027 & 0.017 & 0.672 \\
pro\_sup & 0.002 & 0.001 & 0.246 & 0.002 & 0.001 & 0.255 & 0.002 & 0.001 & 0.193 & 0.002 & 0.001 & 0.225 & 0.002 & 0.001 & 0.304 & 0.002 & 0.001 & 0.084 \\
wrist\_flex & 0.003 & 0.002 & 0.727 & 0.003 & 0.002 & 0.735 & 0.003 & 0.002 & 0.707 & 0.003 & 0.002 & 0.719 & 0.003 & 0.002 & 0.752 & 0.003 & 0.002 & 0.630 \\
wrist\_dev & 0.003 & 0.002 & 0.589 & 0.003 & 0.002 & 0.562 & 0.004 & 0.002 & 0.529 & 0.003 & 0.002 & 0.551 & 0.003 & 0.002 & 0.615 & 0.004 & 0.002 & 0.459 \\
\midrule
\textbf{Upper mean} & 0.008 & 0.004 & 0.579 & 0.008 & 0.004 & 0.575 & 0.008 & 0.004 & 0.545 & 0.008 & 0.004 & 0.561 & 0.007 & 0.004 & 0.612 & 0.009 & 0.006 & 0.461 \\
\midrule
\multicolumn{19}{c}{\textit{Lower body}} \\
\midrule
hip\_flex & 0.687 & 0.405 & 0.648 & 0.687 & 0.403 & 0.637 & 0.689 & 0.410 & 0.637 & 0.683 & 0.402 & 0.655 & 0.680 & 0.392 & 0.658 & 0.787 & 0.506 & 0.580 \\
hip\_add & 0.576 & 0.302 & 0.588 & 0.575 & 0.303 & 0.586 & 0.598 & 0.330 & 0.553 & 0.592 & 0.313 & 0.558 & 0.577 & 0.294 & 0.589 & 0.621 & 0.348 & 0.515 \\
hip\_rot & 0.425 & 0.140 & 0.258 & 0.424 & 0.138 & 0.259 & 0.431 & 0.151 & 0.243 & 0.426 & 0.143 & 0.257 & 0.425 & 0.138 & 0.267 & 0.445 & 0.174 & 0.176 \\
knee & 0.593 & 0.337 & 0.780 & 0.595 & 0.342 & 0.767 & 0.605 & 0.349 & 0.779 & 0.614 & 0.354 & 0.770 & 0.595 & 0.327 & 0.785 & 0.727 & 0.428 & 0.729 \\
ankle & 0.541 & 0.288 & 0.806 & 0.544 & 0.292 & 0.802 & 0.550 & 0.305 & 0.799 & 0.544 & 0.295 & 0.805 & 0.535 & 0.268 & 0.812 & 0.646 & 0.386 & 0.732 \\
mtp & 0.001 & 0.001 & 0.837 & 0.001 & 0.001 & 0.812 & 0.001 & 0.001 & 0.816 & 0.001 & 0.001 & 0.823 & 0.001 & 0.001 & 0.849 & 0.001 & 0.001 & 0.704 \\
\midrule
\textbf{Lower mean} & 0.470 & 0.245 & 0.653 & 0.471 & 0.246 & 0.644 & 0.479 & 0.258 & 0.637 & 0.477 & 0.251 & 0.645 & 0.469 & 0.236 & 0.660 & 0.538 & 0.307 & 0.572 \\
\midrule
\textbf{All joints} & 0.285 & 0.149 & 0.623 & 0.286 & 0.149 & 0.616 & 0.291 & 0.156 & 0.600 & 0.289 & 0.153 & 0.611 & 0.284 & 0.143 & 0.641 & 0.326 & 0.186 & 0.528 \\
\bottomrule
\end{tabular}%
}
\\[2pt]
\begin{flushleft}\footnotesize R: RMSE (Nm/kg)$\downarrow$ \quad M: MAE (Nm/kg)$\downarrow$ \quad P: PCC$\uparrow$\end{flushleft}
\end{table*}

\section{Input Feature Ablation}\label{supp:input_ablation}

Here we ablate the parametric body channel by training a Transformer on poses alone and on poses paired with $\bm{\beta}$, mass, and height, all under an identical training protocol. Table~\ref{tab:supp_input_ablation} reports the result. Adding the parametric body inputs reduces the All joints RMSE from $0.307$ to $0.291$\,Nm/kg ($-5\%$) and the MAE from $0.164$ to $0.156$\,Nm/kg ($-5\%$), with a small gain in Pearson correlation ($+1\%$). The improvement comes from the lower body, where the body shape carries information about segment masses and lengths that drives the inverse-dynamics torques. Upper-body errors are essentially unchanged, since their numerical magnitude is small ($\sim\!10^{-2}$\,Nm/kg) and the parametric body channel adds little new signal at that scale.

\begin{table}[h]
\centering
\small
\caption{Input feature ablation on the Transformer baseline. Adding the parametric body inputs (SKEL shape $\bm{\beta}$, mass, height) on top of pose-only inputs reduces All joints torque error by about $5\%$.}
\label{tab:supp_input_ablation}
\setlength{\tabcolsep}{6pt}
\resizebox{\textwidth}{!}{%
\begin{tabular}{l|cc|ccc|ccc|ccc}
\toprule
\multirow{2}{*}{\textbf{Condition}} & \multicolumn{2}{c|}{\textbf{Inputs}} & \multicolumn{3}{c|}{\textbf{Upper mean}} & \multicolumn{3}{c|}{\textbf{Lower mean}} & \multicolumn{3}{c}{\textbf{All joints}} \\
\cmidrule(lr){2-3} \cmidrule(lr){4-6} \cmidrule(lr){7-9} \cmidrule(lr){10-12}
 & poses & shape
 & R$\downarrow$ & M$\downarrow$ & P$\uparrow$
 & R$\downarrow$ & M$\downarrow$ & P$\uparrow$
 & R$\downarrow$ & M$\downarrow$ & P$\uparrow$ \\
\midrule
\textbf{(A) poses only}
  & \checkmark &
  & 0.009 & 0.005 & 0.473
  & 0.506 & 0.270 & 0.608
  & 0.307 & 0.164 & 0.554 \\
\textbf{(B) +parametric body}
  & \checkmark & \checkmark
  & 0.008 & 0.005 & 0.484
  & 0.480 & 0.256 & 0.607
  & 0.291\,\textcolor{brightpurple}{($-5\%$)}
  & 0.156\,\textcolor{brightpurple}{($-5\%$)}
  & 0.558\,\textcolor{brightpurple}{($+1\%$)} \\
\bottomrule
\end{tabular}%
}
\\[2pt]
\begin{flushleft}\footnotesize R: RMSE (Nm/kg) \quad M: MAE (Nm/kg) \quad P: PCC.\end{flushleft}
\end{table}

\bibliographystyle{splncs04}
\bibliography{main}